\documentclass[a4paper,11pt,fleqn]{article}
\usepackage{PRIMEarxiv}
\usepackage{amsmath,mathrsfs,amssymb}
\usepackage[linesnumbered,ruled,vlined]{algorithm2e}
\usepackage[authoryear]{natbib}
\usepackage[utf8]{inputenc} 
\usepackage[T1]{fontenc}    
\usepackage{diagbox}
\usepackage{caption, subcaption}
\usepackage{multicol, multirow}
\usepackage{mathtools}
\usepackage{hyperref}       
\usepackage{url}            
\usepackage{amsfonts}       
\usepackage{nicefrac}       
\usepackage{microtype}      
\usepackage{fancyhdr}       
\usepackage{booktabs,threeparttable}
\usepackage{adjustbox}
\usepackage{stfloats}
\usepackage{tabularx}
\usepackage{graphicx}       
\graphicspath{{media/}}     

\def\tsc#1{\csdef{#1}{\textsc{\lowercase{#1}}\xspace}}
\tsc{WGM}
\tsc{QE}
\tsc{EP}
\tsc{PMS}
\tsc{BEC}
\tsc{DE}

\makeatletter
\newcommand*\rel@kern[1]{\kern#1\dimexpr\macc@kerna}
\newcommand*\widebar[1]{%
  \begingroup
  \def\mathaccent##1##2{%
    \rel@kern{0.8}%
    \overline{\rel@kern{-0.8}\macc@nucleus\rel@kern{0.2}}%
    \rel@kern{-0.2}%
  }%
  \macc@depth\@ne
  \let\math@bgroup\@empty \let\math@egroup\macc@set@skewchar
  \mathsurround\z@ \frozen@everymath{\mathgroup\macc@group\relax}%
  \macc@set@skewchar\relax
  \let\mathaccentV\macc@nested@a
  \macc@nested@a\relax111{#1}%
  \endgroup
}
\makeatother

\newcommand{\SubItem}[1]{
    {\setlength\itemindent{15pt} \item[-] #1}
}

\newtheorem{condition}{Condition}

\begin{document}

\fancyhead[LO]{On Solving CEOP with Overlapped Neighborhoods}
\title{On Solving Close Enough Orienteering Problems with Overlapped Neighborhoods}
\author{Qiuchen Qian, Yanran Wang, David Boyle}
\maketitle

\begin{abstract}
The Close Enough Traveling Salesman Problem (CETSP) is a well-known variant of the classic Traveling Salesman Problem whereby the agent may complete its mission at any point within a target neighborhood. Heuristics based on overlapped neighborhoods, known as Steiner Zones (SZ), have gained attention in addressing CETSPs. While SZs offer effective approximations to the original graph, their inherent overlap imposes constraints on the search space, potentially conflicting with global optimization objectives. Here we show how such limitations can be converted into advantages in the Close Enough Orienteering Problem (CEOP) by aggregating prizes across overlapped neighborhoods. We further extend the classic CEOP with Non-uniform Neighborhoods (CEOP-$\mathcal{N}$) by introducing non-uniform cost considerations for prize collection. To tackle \underline{\textbf{C}}EOP (and CEOP-$\mathcal{N}$), we develop a new approach featuring a \underline{\textbf{Ra}}ndomized \underline{\textbf{S}}teiner \underline{\textbf{Z}}on\underline{\textbf{e}} Discretization (RSZD) scheme coupled with a hybrid algorithm based on Particle Swarm Optimization (PSO) and \underline{\textbf{Ant}} Colony \underline{\textbf{S}}ystem (ACS) - CRaSZe-AntS. The RSZD scheme identifies sub-regions for PSO exploration, and ACS determines the discrete visiting sequence. We evaluate the RSZD's discretization performance on CEOP instances derived from established CETSP instances and compare CRaSZe-AntS against the most relevant state-of-the-art heuristic focused on single-neighborhood optimization for CEOP instances. We also compare the performance of the interior search within SZs and the boundary search on individual neighborhoods in the context of CEOP-$\mathcal{N}$. Our experimental results show that CRaSZe-AntS can yield comparable solution quality with significantly reduced computation time compared to the single neighborhood strategy, where we observe an averaged $140.44\,\%$ increase in prize collection and $55.18\,\%$ reduction of algorithm execution time. CRaSZe-AntS is thus highly effective in solving emerging CEOP-$\mathcal{N}$, examples of which include truck-and-drone delivery scenarios.
\end{abstract}

\keywords{Metaheuristics \and Close enough orienteering problem \and Steiner zone discretization \and Non-uniform neighborhood \and Truck-and-drone problems}

\section{Introduction} \label{sec:introduction}
The Close Enough Traveling Salesman Problem (CETSP) \citep{gulczynski2006close}, a generalized extension of the classic Traveling Salesman Problem (TSP), has attracted substantial attention in the recent research literature. CETSP aims to determine the most efficient path originating from a designated depot, traversing all service regions, and ultimately returning to the initial depot. One significant application of CETSP is mission planning for Unmanned Aerial Vehicles (UAV) due to their important characteristic of having the ability to effectively execute missions within predefined spatial constraints \citep{gao2020routing, cariou2023evolutionary}. CETSP is extensively addressed in the literature, employing exact methods \citep{coutinho2016branch, ha2012exact} and heuristic approaches \citep{yang2018double, wang2019steiner, carrabs2020adaptive} for solution finding. A notable heuristic proposed by \cite{carrabs2020adaptive}, namely (lb/ub)Alg, effectively addresses CETSP's discretization by tackling the Generalized Traveling Salesman Problem (GTSP) and refining waypoints through Second-Order Cone Programming (SOCP). Despite the foundational significance of CETSP in the community, it is essential to acknowledge three key limitations of the CETSP model in this context — 1. the requirement for the agent to commence and terminate its mission at the same depot presents practical challenges across various scenarios. For example, it conflicts with the dynamic nature of UAV-related mission planning, which often demands real-time adjustments in response to complex environmental changes \citep{evers2014online, bayliss2020learnheuristic, qian2022practical}; 2. TSP is essentially an unconstrained optimization problem, while real-world agent operations are invariably subject to diverse constraints such as time and motion capacity; 3. The priority for the agent to visit or serve one target is unlikely to be uniform. For instance, \citet{di2023generalized} proposed the Generalized CETSP with extra `profit' assigned for each target node for RFID meter reading systems, representing the probability of correct reading. \citet{qian2021optimal} demonstrated the need for assigning varying priorities to sensor nodes to prevent data loss in a UAV-assisted wireless recharging scheduling scenario. 

Therefore, we introduce the Close Enough Orienteering Problem (CEOP) to handle the above limitations of CETSP. The CEOP model seeks to identify an optimal path that initiates at a specified start depot and returns to a different end depot (Limitations 1), maximizing the collected prize (Limitations 3) without violating the given budget constraint (Limitations 2). Though approaches to CEOP have been widely discussed in the literature, such as the heuristics \citep{pvenivcka2017dubins, vstefanikova2020greedy} and machine learning approaches \citep{faigl2018gsoa, deckerova2020hopfield}, there remains a notable gap in research concerning the intricacies of solving CEOP within overlapped neighborhoods. In this context, \textit{overlapped neighborhoods} refers to scenarios where the agent can collect prizes from multiple target nodes concurrently. This concept introduces a novel perspective, turning the limitations of overlapped neighborhoods, a.k.a., Steiner Zones (SZ) employed in CETSP instances \citep{gulczynski2006close} (see $\S$\ref{sec:related-work} for more details), into advantages by leveraging prize-based attributes unique to CEOP. Specifically, CEOP instances can be reduced to Set Orienteering Problem (SOP) \citep{archetti2018set} or even OP instances by aggregating prizes across overlapped neighborhoods. Additionally, the practicality of this prize aggregation often requires the consideration of non-linear cost functions. For instance, the Orthogonal Frequency Division Multiple Access technique allows the UAV to establish communication links with multiple sensor nodes \citep{chen2021data}, but the data transmission efficiency heavily depends on the antenna orientation and relative positioning. Similarly, within the framework of a Truck-and-Drone Delivery Problem (TDDP), multiple UAVs can be deployed to deliver packages to nearby customers in parallel while the truck's idle time depends on its stop position and UAV operation time \citep{boysen2018drone}. Therefore, to further extend the generality of CEOP, we propose a novel model, CEOP with non-uniform neighborhoods, abbreviated as CEOP-$\mathcal{N}$.

This paper starts from the original CEOP formulation, investigating the geometric attributes of (overlapped) neighborhoods, and subsequently delves into their implications on CEOP-$\mathcal{N}$ when involving non-uniform cost functions. We also discuss the sub-problem formulation from the discretization of the CEOP (i.e., SOP). It is essential to underscore that the discretization of CEOP-$\mathcal{N}$ translates to a standard OP only because CEOP-$\mathcal{N}$ requires the interior search within overlapped neighborhoods. Our key contributions can be summarized as follows:
\begin{itemize}
    \item We formulate a new optimization problem, Close Enough Orienteering Problem with non-uniform neighborhoods (CEOP-$\mathcal{N}$). The objective is to maximize the collected prizes while adhering to the joint constraints of the prize-collecting costs within a neighborhood and travel expenses incurred by the agent. We support CEOP-$\mathcal{N}$ with a straightforward Truck-and-drone Delivery Problem (TDDP) example.
    \item{We develop the Randomized Steiner Zone Discretization (RSZD) scheme to discretize the original CEOP and CEOP-$\mathcal{N}$ graph with Steiner Zones, characterized by two or more circular arcs. Our empirical findings indicate that RSZD outperforms the Internal Point Discretization (IPD) scheme \citep{carrabs2020adaptive}. For instance, consider the results of solving a discrete instance of CEOP (\textit{bubbles5}) as the example. When utilizing RSZD's layout as opposed to IPD's, the solver achieves notable enhancements in both computational efficiency (approximately 30 times faster) and solution quality (a prize gain of $24.32\,\%$).}
    \item{We demonstrate the significance of involving the SZ-based strategy in solving \underline{C}EOP with overlapped neighborhoods and (non-)uniform prize-collecting cost functions. We develop an efficient and reliable algorithm named CRaSZe-AntS, utilizing the \underline{Ra}ndomized \underline{S}teiner \underline{Z}on\underline{e} Discretization scheme and the \underline{Ant} Colony \underline{S}ystem (ACS). CRaSZe-AntS is constituted of different solvers and local search operators when addressing the uniform and non-uniform scenarios, specifically:}
        \SubItem{For uniform CEOP, CRaSZe-AntS employs an efficient local search operator named arc search algorithm to optimize the waypoint positions generated by RSZD. We compare the performance of CRaSZe-AntS with the most relevant benchmark algorithm, namely Greedy Randomized Adaptive Search Procedure (GRASP) \citep{vstefanikova2020greedy}. Our experimental findings demonstrate that CRaSZe-AntS can yield solutions with comparable quality within remarkably reduced computation time ($-40.6\,\%$ on average) across large-scale CEOP instances.}
        \SubItem{For non-uniform CEOP, i.e., CEOP-$\mathcal{N}$, we examine the interior space of overlapped neighborhoods. CRaSZe-AntS combines Particle Swarm Optimization (PSO), which searches the optimal waypoints' positions within overlapped neighborhoods, and an Inherited Ant Colony System (IACS) that converges quicker than classic ACS. We compare the performance of CRaSZe-AntS with the most relevant benchmark algorithm, namely Hybrid Algorithm (HA) \citep{yang2018double}. Our experimental results show that CRaSZe-AntS can significantly reduce the computation time ($-55.18\,\%$ on average) and improve solution quality ($140.44\,\%$ on average) across all instances tested, compared to the strategy that considers single circle only.}
    \item{Software implementation and experimental results of CRaSZe-AntS are made openly available to the community\footnote[1]{\url{https://github.com/sysal-bruce-publication/CRaSZe-AntS.git} \label{fn-github}}.}
\end{itemize}

\section{Related Work} \label{sec:related-work}
This section provides an overview of algorithms applied to address problems associated with `Close-Enough' scenarios. These strategies can be broadly categorized into SZ strategies and single-neighborhood strategies.

\citet{gulczynski2006close} initially discretized the CETSP instance with \textit{super nodes} assigned to individual target nodes. Their heuristics comprise three primary phases - (1) generation of feasible super nodes; (2) construction of a solution path based on the super node set by solving a Generalized Traveling Salesman Problem (GTSP) \citep{henrylab1969record}; (3) local improvements made to the super nodes within the solution path. A noteworthy heuristic known as the Steiner Zone Heuristic (SZH) plays a central role. Specifically, SZH constructs SZs, which represent the intersection of circles (a formal definition is given in \S \ref{sec:sz-def}), and examines the super nodes' feasibility by checking their intersections with all circles covered by the SZ. Though SZH has been shown to yield solutions of superior quality compared to other heuristics presented in \citep{gulczynski2006close}, super nodes were arbitrarily selected within all SZs to form the feasible set. \citet{mennell2009heuristics} further refined SZH by introducing limited Steiner Zone degrees (defined in \S \ref{sec:sz-def}). For instance, the minimum-angle strategy restricts the insertion circle with a minimum insertion angle threshold. Following the framework in \citep{gulczynski2006close}, \citet{mennell2009heuristics} designates the intersection chord's midpoint as the representative point for the SZ. The path is then improved using another heuristic, known as $IP_{phIII}$, that iteratively improves the position of a representative point within a certain arc range. However, SZH exhibits a slow convergence rate because it iteratively narrows the search range of the arc across all SZs. \citet{wang2019steiner} proposed the Steiner Zone Variable Neighborhood Search (SZVNS) heuristic, which replaces the path improvement phase in \citep{mennell2009heuristics} with a bisection algorithm. The algorithm identifies the waypoint $B$ on the arc of a SZ (or within a SZ if $B$ falls on line segment $\overline{AB}$) that minimizes the distance of line segments $\overline{AB}+\overline{BC}$, where $A, C$ represent the previous and next waypoints in the path. SZVNS has demonstrated comparable solution quality to SZH while significantly reducing computational time. Despite their ability to swiftly generate (near-)optimal solutions (compared to the single neighborhood strategy), SZ-based strategies have limitations in addressing CETSP. The SZ strategy can be seen as an approximation of the original graph, but constructing an optimal SZ layout for the input graph remains challenging, as highlighted in \citep{mennell2009heuristics, wang2019steiner}. For instance, \citet{mennell2009heuristics} have introduced multiple metrics to regulate the addition of circles to a SZ. SZVNS attempts to iteratively improve layout quality by regenerating three candidate SZ layouts. Additionally, SZs confine the search space of multiple neighborhoods into smaller convex regions, limiting solution quality. On the other hand, optimizing the path's waypoints on single neighborhoods can circumvent these limitations of SZs in CETSP. \citet{carrabs2017novel} and \citet{carrabs2020adaptive} present a novel discretization scheme that sets discrete points for each target circle. Once the path sequence is determined, the optimal solution is then found using the IBM ILOG CPLEX solver by solving the formulated second-order cone programming model. \citet{yang2018double} adopted Particle Swarm Optimization (PSO) with a Boundary Encoding (BE) scheme for optimizing waypoint positions on the boundary of arbitrary neighborhoods and a Genetic Algorithm (GA) for optimizing the path sequence and evaluating the fitness of the particles. Similarly, \citet{di2022genetic} employed a reverse logic approach, using GA and random sampling to generate a feasible solution and then optimizing continuous positions with the IBM ILOG CPLEX solver. 

Though SZ-based strategies have been widely employed in addressing CETSP, their application to the CEOP and its variants remains unexplored in the existing literature. A notable implementation of a single-neighborhood strategy for CEOP is GRASP, as detailed by \cite{vstefanikova2020greedy}, which utilizes a local search based on the geometric relationship between the circle and the line segment. However, GRASP's efficacy diminishes with increasing problem scales, attributed to its reliance on an `elite' selection strategy (discussed in $\S$ \ref{sec:result-ceop}). Adapting SZ-based strategies for CEOP can effectively reduce the problem complexities by aggregating prizes of overlapped target circles into a unified SZ. This approach mirrors the discretization process in CETSP, allowing CEOP to be simplified to a SOP \citep{archetti2018set}. SOPs can then be efficiently tackled using metaheuristics or exact methods. For example, \cite{pvenivcka2019variable} introduced a Variable Neighborhood Search (VNS) metaheuristic alongside an Integer Linear Programming formulation for SOP. \cite{archetti2024new} proposed another efficient formulation and a branch and bound algorithm for SOP. Leveraging the solutions derived from SOP, local search operators can then be employed to enhance the solution quality of CEOP.

It is worth noting that real-world applications often require more flexibility than what CETSP (or CEOP) typically offers - an assumption that the agent arriving somewhere close to targets is sufficient to complete the mission. For example, \citet{cariou2023evolutionary} examined CETSP solutions within the context of communication intersections. Given that antenna radiation patterns are not perfectly isotropic, the original CETSP model would yield suboptimal solutions unless penalties are introduced based on the radio wave strength within the circle's interior. \citet{gao2020routing} introduced the Risk and Reward Asset Routing Problem (R$^2$ARP), which demands a delicate balance between the risk and reward within the neighborhood. \citet{mantzaras2017optimization} proposed an optimization model for siting, collecting, and transferring Infectious Medical Waste (IMW). Each node's location and waste production rate (e.g., hospital, health center, clinic, etc.) are critical factors for the cost function and thus lead to a non-uniform neighborhood. These examples highlight the advantages of considering the overlap between target neighborhoods and corresponding nonlinear cost functions. Evidently, the single neighborhood strategy is inefficient in such a context. Furthermore, searching the interior of target (overlapped) neighborhoods is necessary. However, at the time of writing this paper, boundary searching for neighborhoods remained the main strategy for addressing problems related to `Close-Enough' scenarios \citep{coutinho2016branch, carrabs2020adaptive, di2022genetic, faigl2019data, di2023generalized}. Consequently, we employ the SZ strategy to establish the boundary conditions for the interior search phase during the waypoint optimization.
\vspace*{-0.4cm}

\section{Problem Formulation} \label{sec:problem-formulation}
In this section, we present the formulation of CEOP incorporating SZ constraints within the context of both uniform and non-uniform neighborhoods. Uniform and non-uniform neighborhoods are characterized by the nature of their cost functions for prize collection. For ease of reference, we employ regular font abbreviations, such as CEOP and CEOP-$\mathcal{N}$, to denote the problem. We use the bold font \textbf{abbreviation} enclosed in \textbf{brackets} to denote the mathematical formulations provided in this section, e.g., $\textbf{(CEOP)}$ and $\textbf{(CEOP-}\boldsymbol{\mathcal{N}}\textbf{)}$. We provide an overview of the notation used in the following formulations in Table \ref{tab:notation}.
\begin{figure}[!b]
    \centering	
    \includegraphics[width=0.45\textwidth]{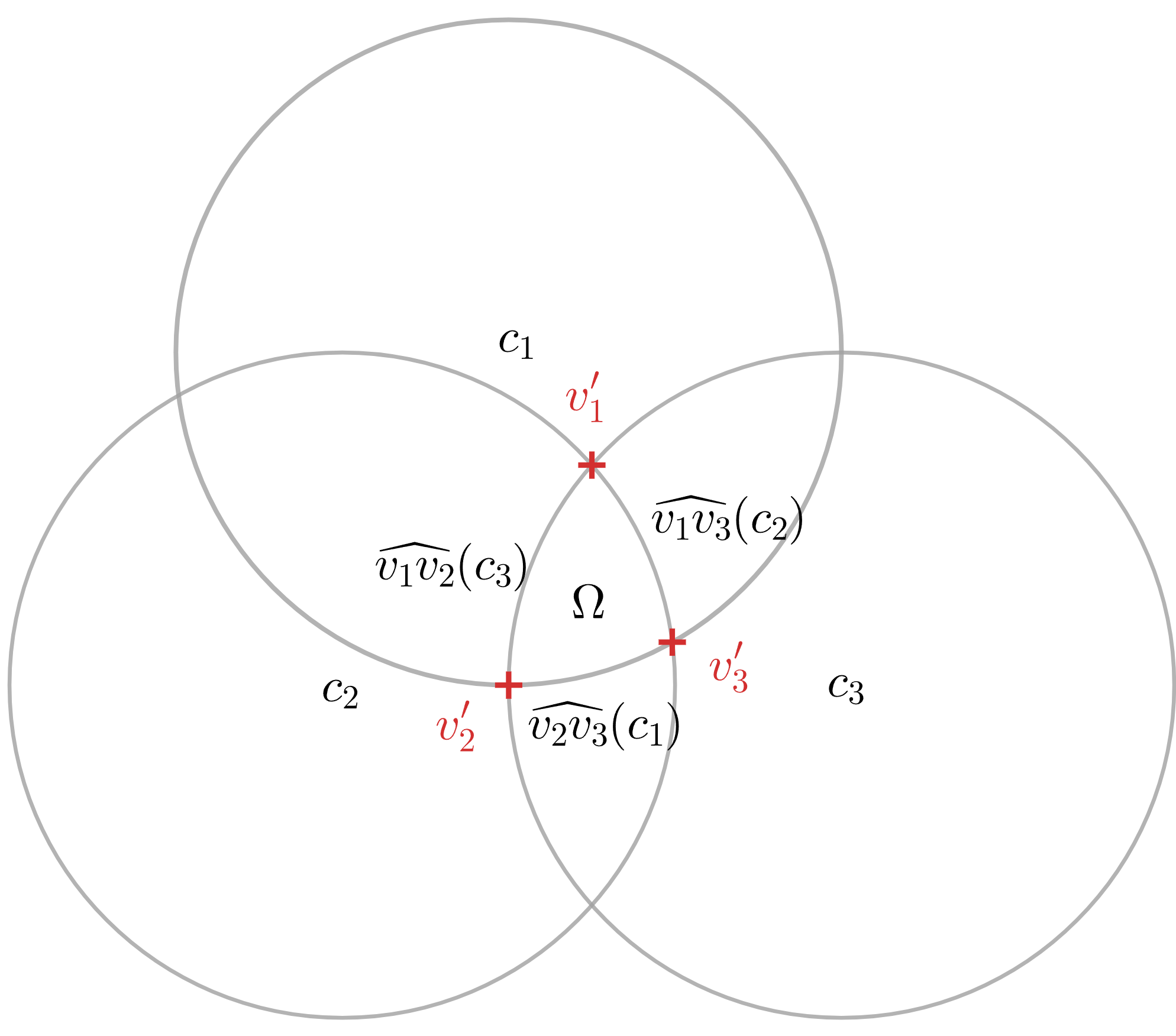}
    \caption{A Steiner zone of degree three, $|\Omega| = 3$, formed by three circular arcs $\widehat{v'_{1} v'_{2}}(c_3)$, $\widehat{v'_{1} v'_{3}}(c_2)$, and $\widehat{v'_{2} v'_{3}}(c_1)$. The red cross points refer to SZ vertices $v'_1, v'_2, v'_3$.}
    \label{fig:sz3}
\end{figure}

\begin{table}[!t] 
\centering
\caption{Notation used in \S 3 Problem Formulation.} \label{tab:notation}
\begin{tabularx}{\textwidth}{ll}
    \toprule \textit{Geometry related parameters} \\ \midrule
    $c_i$,\: $v_i$,\: $r_i$ & Target circle with index $i$, and its corresponding circle center $v_i$ and radius $r_i$. \\
    $\Omega_i$ & The Steiner Zone with index $i$. \\
    $|\Omega_i|$ & The number of circles covered by the Steiner Zone $\Omega_i$. \\
    $v'(c_i)$, $v'(\Omega_j)$ & Any waypoint (discrete point) located in the circle $c_i$ or the Steiner Zone $\Omega_j$. \\
    $v'_{ctr}(\Omega_i)$ & The center of the Steiner Zone $\Omega_i$. \\
    $(x_i, y_i)$, $(x'_j, y'_j)$ & Two-dimensional coordinate of the circle center $v_i$ or the waypoint $v'_j$. \\ 
    $\widehat{v'_i \: v'_j}(c_k)$ & The circular arc on the circle $c_k$ that starts from the vertex $v'_i$ to the vertex $v'_j$. \\
    \midrule \textit{Indices and constants} \\ \midrule
    Subscript $i, j, k, l$ & The indices used for distinguishing target circles, waypoints, Steiner Zones, etc. \\
    Subscript $0$, $N+1$ & The indices used to represent the start and end depot. \\
    Subscript $0$, $K+1$ & The indices used to represent the start and end depot. They are only used in the \\
     & \;  Set Orienteering Problem (SOP) because SOP's inputs are Steiner Zone vertices. \\
    $N$ & The number of target circles. \\
    $N_{\max}^{|\Omega|}$ & The maximum number of circles can be included in a Steiner Zone. \\
    $M$ & The number of Steiner Zones, obtained by solving the Steiner Zone Search Problem. \\
    $K$ & The number of waypoints, $K = \sum_{i=1}^{M} |\Omega_i|$. \\
    \midrule \textit{Sets} \\ \midrule
    $\mathbf{G} = \big\{ \mathbf{C}, \mathbf{E} \big\}$ & The complete graph with target circle set $\mathbf{C} = \big\{ c_1, ..., c_N \big\}$ and edge set $\mathbf{E} = \big\{ e_{ij}, ... \big\}$. \\
    $\mathbf{Z}$ & The set of Steiner Zones $\mathbf{Z} = \big\{ \Omega_1, ..., \Omega_M \big\}$. \\
    $\mathbf{V}$ & The set of Steiner Zone vertices $\mathbf{V} = \big\{ v'_1, ..., v'_K \big\}$. \\ 
    \midrule \textit{Problem related parameters} \\ \midrule
    $\mathscr{C}_{f_1} (v' \:|\: c_j)$ & The general prize-collecting cost of the circle $c_j$ at a waypoint $v'$, defined by a \\ 
     & \; continuous function $f_1$. \\
    $\mathscr{C}_t (v' \:|\: \Omega_j)$ & The prize-collecting cost of a Steiner Zone $\Omega_j$ at a waypoint $v'$, defined in time metric. \\
    $\mathscr{C}_{f_2} (v'_i,\: v'_j)$ & The general traveling cost from $v'_i$ to $v'_j$, defined by a continuous function $f_2$. \\
    $\mathscr{C}_d (v'_i,\: v'_j)$, $\mathscr{C}_t (v'_i,\: v'_j)$ & The traveling cost from $v'_i$ to $v'_j$, defined in the distance or time metric. \\
    $\mathscr{P}(c_i)$, $\mathscr{P}(\Omega_j)$ & The prize of a circle $c_i$ or a Steiner Zone $\Omega_j$. \\
    $\mathscr{B}_d$, $\mathscr{B}_t$, $\mathscr{B}_f$ & The budget constraint of a path, defined in the distance, time, or general form metric. \\
    $\mathcal{V}_{\text{drone}}$, $\mathcal{V}_{\text{truck}}$ & The average flight speed of the delivery UAV or the average truck speed. \\
    $\lambda$ & The flight efficiency (in percentage) of the delivery UAV with a weight load. \\
    $t_{\text{serv}}$ & The service time of the UAV, including landing, handing over package, and takeoff. \\ 
    \bottomrule
\end{tabularx}
\end{table}

\subsection{Steiner Zone Search Problem (SZSP)} \label{sec:sz-def}
We first introduce the definition of SZs, as it forms the foundation for understanding overlapped neighborhoods. A SZ, denoted by $\Omega$, is a convex region enclosed by one or more circular arcs. The \textit{degree} of a SZ refers to the number of circles it covers. For example, we use $|\Omega| = 3$ to represent a SZ formed by three circles. Given two intersect points ($v'_1$ and $v'_2$) on the circumference of two circles ($c_1$ and $c_2$), the \textit{circular arc} on $c_1$ (from $v'_1$ to $v'_2$) is denoted as $\widehat{v'_{1} v'_{2}}(c_1)$. We use $v' (\Omega_k)$ with coordinate $\big(x'(\Omega_k), y'(\Omega_k)\big)$ to represent any \textit{waypoint} located in the SZ $\Omega_k$. The vertices of a SZ, also belonging to special waypoints, correspond to the starting and ending points of its constituent circular arcs. For consistency, we still use $v'$ to denote a vertex of the SZ. Fig.~\ref{fig:sz3} demonstrates a SZ of degree three, which is formed by three circular arcs $\widehat{v'_{1} v'_{2}}(c_3)$, $\widehat{v'_{1} v'_{3}}(c_2)$, and $\widehat{v'_{2} v'_{3}}(c_1)$. This SZ has three vertices $v'_1, v'_2, v'_3$. Moreover, the coordinate of a SZ's \textit{center} $v'_{\text{ctr}} (\Omega_j)$ is defined as:
\vspace*{-0.1cm}
\begin{equation}
    \Big( x'_{\text{ctr}}(\Omega_j),\: y'_{\text{ctr}}(\Omega_j) \Big) = \begin{dcases}
        \text{circle center}\; , & \text{if } |\Omega| = 1\\
        \Big( \dfrac{\sum_{v'_i \in\:\Omega_j} x'_i}{|\Omega_j|}, \: \dfrac{\sum_{v'_i \in\:\Omega_j} y'_i}{|\Omega_j|} \Big)\; , & \text{otherwise}
    \end{dcases}
\end{equation}
In other words, the coordinate of a SZ's center is the average of all SZ vertices. The conditions for adding an external circle $c_i$ to an existing SZ $\Omega_j$ are as follows:
\begin{condition} \label{prop:suff}
    The circle $c_i$ must intersect with all other circles $c_k \in \Omega_j$. 
\end{condition}
\begin{condition} \label{prop:nece}
    For each pair of intersecting circles, at least one intersection point must reside within the boundaries of those circles or on their circumferences.
\end{condition}
Note that a circle may also be added to the SZ if it is entirely contained within another larger circle. However, this specific scenario lies beyond the scope of our study, as it necessitates a trivial preprocessing step to accommodate contained circles. Thus, we assume a uniform circle radius for all circles. Given the circle set $\mathbf{C} = \big\{ c_1, ..., c_N \big\}$, the Steiner Zone Search Problem (SZSP) aims at reconstructing a SZ layout $\mathbf{Z} = \big\{ \Omega_1, ..., \Omega_M \big\}$, where $M \leq N$, that covers all circles. With the decision variable $Y_{ij} \in \big\{ 0, 1 \big\}$ for each circle $c_i$ to be added in the SZ $\Omega_j$, the formulation of SZSP is given as follows:
\begin{subequations}
\begin{flalign}
    \label{eq:szsp-obj}    
    \textbf{(SZSP)}\quad\quad &\min~\vert \mathbf{Z} \vert\\
    \label{eq:szsp-one-circle}
    & \sum_{i=1}^{N} \sum_{j=1}^{M} Y_{ij} = 1 &\\  
    \label{eq:szsp-max-degree}
    & \sum_{i=1}^{N} Y_{ij} \leq N^{|\Omega|}_{\max}, ~~j = 1, ..., M &\\
    \label{eq:szsp-prop}
    & Y_{ij} \leq \begin{cases}
        1, & \text{if } c_i \text{ and } \Omega_j \text{ satisfy \textbf{Condition}}~\ref{prop:suff}, \ref{prop:nece} \\
        0, & \text{otherwise}
    \end{cases} &
\end{flalign}
\end{subequations}
The objective function \eqref{eq:szsp-obj} minimizes the number of SZs. Constraint \eqref{eq:szsp-one-circle} enforces each circle can be uniquely added into one specific SZ. Constraint \eqref{eq:szsp-max-degree} restricts the maximum degree of a SZ to $N^{|\Omega|}_{\max}$. Constraint \eqref{eq:szsp-prop} requires a circle to be added to a SZ must satisfy \textbf{Condition} \ref{prop:suff} and \ref{prop:nece}. 

\subsection{CEOP and uniform overlapped neighborhoods}
Let $\mathbf{G} = \big\{ \mathbf{C}, \mathbf{E} \big\}$ be a complete graph with a set of $N$ target circles $\mathbf{C} = \big\{ c_1, ..., c_N \big\}$ and a corresponding edge set $\mathbf{E} = \big\{ e_{ij}, ... \big\}$. Each target circle $c_i$ is characterized by its center $v_i$, radius $r_i$, and prize $\mathscr{P}(c_i)$. Each edge $e_{ij}$ represents the connection between two circle centers $v_i$ and $v_j$. A target circle's prize $\mathscr{P}(c_i)$ can be collected at any waypoint $v'$ located within the circle (or on its boundary). We define the distance cost between two circle centers $v_i$ and $v_j$ as Euclidean distance, i.e., $\mathscr{C}_d\big( v_i,\: v_j \big) = \sqrt{(x_i - x_j)^2 + (y_i - y_j)^2}$. We denote the starting depot by $v'_0$ and the ending depot by $v'_{N+1}$ with two-dimensional coordinates $(x'_0,\: y'_0)$ and $(x'_{N+1},\: y'_{N+1})$, respectively. Note that $\mathscr{P}(v'_0) = \mathscr{P}(v'_{N+1}) = 0$. A feasible path of CEOP starts at $v'_0$, collects as much prize as possible, and ends at $v'_{N+1}$, without violating the given distance budget $\mathscr{B}_d$ constraint. Because the CEOP model owns decision variables $X_{ij} \in \big\{ 0, 1 \big\}$ (decide the visitation for all target circles and depots), alongside continuous waypoint coordinates, and quadratic constraints from circular geometry, it can be formulated as a Mixed Integer Nonlinear Programming (MINLP) problem:
\begin{subequations}
\begin{flalign}
    \label{eq:ceop-obj}
    \textbf{(CEOP)}\quad & \max~\sum_{i=0}^N \sum_{j=1}^{N} \mathscr{P}(c_j) \cdot X_{ij}&\\
    \label{eq:ceop-start-end}
    \textbf{s.t.} \quad & \sum_{j=1}^{N+1} X_{0 \: j} = \sum_{i=0}^N X_{i \: N+1} = 1&\\
    \label{eq:ceop-one-visit}
    &\sum_{i=0}^{N+1} X_{ik} = \sum_{j=0}^{N+1} X_{kj} \leq 1, & k = 2, ..., N&\\
    \label{eq:ceop-subtour}
    &\sum_{c_i \in\: \mathbf{S}}\: \sum_{c_j \in\: \mathbf{S}} X_{ij} \leq |\mathbf{S}| - 1, & \forall\: \mathbf{S} \subset \mathbf{C},\; |\mathbf{S}| \geq 3&\\
    \label{eq:ceop-budget}
    &\sum_{i=0}^{N} \: \sum_{j=1}^{N+1} \mathscr{C}_d\big( v'_i,\: v'_j \big) \cdot X_{ij}\: \leq \:\mathscr{B}_d&\\
    \label{eq:ceop-radius}
    &\mathscr{C}_d\big( v_i,\: v'_i \big) \leq r_i, & i = 1, ..., N&
\end{flalign}
\end{subequations}
The objective function \eqref{eq:ceop-obj} maximizes the collected prizes of all visited target circles. Constraint \eqref{eq:ceop-start-end} ensures the path starts at depot $v_0$ and ends at depot $v_{N+1}$. Constraint \eqref{eq:ceop-one-visit} provides the connectivity of the path and enforces that one target circle is visited at most once. Constraint \eqref{eq:ceop-subtour} prevents subtours. Constraint \eqref{eq:ceop-budget} implies the path distance cost must not exceed the given distance budget $\mathscr{B}_d$. Here we exclude $N+1$ for index $i$ and exclude $0$ for index $j$ because $\sum_{j=0}^{N+1} X_{N+1\: j} = 0$ and $\sum_{i=0}^{N+1} X_{i\: 0} = 0$. Constraint \eqref{eq:ceop-radius} restricts that the waypoint of a target circle must be located within its circular neighborhood or on its boundary. Additionally, by introducing overlapped neighborhoods into the above CEOP formulation, we can give additional constraints from SZs:
\begin{subequations}
\begin{flalign}
    \label{eq:sz-constraint}
     & v' \in \Omega_k \iff \mathscr{C}_d \big( v', v_j \big) \leq r_j, \quad\quad & \forall\: c_j \in \Omega_k,\; k=1,..., M&\\ 
     \label{eq:sz-prize}
     & \mathscr{P}(\Omega_k) = \sum_{c_j\in\:\Omega_k} \mathscr{P}(c_j), \quad\quad & k=1,..., M&
\end{flalign}
\end{subequations} 
Constraint \eqref{eq:sz-constraint} ensures that any waypoint located in a SZ $\Omega_j$ also locates within all SZ-covered circles, and vice versa. Constraint \eqref{eq:sz-prize} defines the prize of a SZ as the prize sum of all SZ-covered circles.

\subsubsection{Set Orienteering Problem (SOP)}
We formalize the SOP as an approximation to \textbf{(CEOP)} based on the vertices of constructed SZs. Given a set of $M$ Steiner Zones $\mathbf{Z} = \big\{ \Omega_1, ..., \Omega_M \big\}$, there are $K = \sum_{i=1}^{M} |\Omega_i|$ SZ vertices. A complete graph $\mathbf{G} = \big\{ \mathbf{V}, \mathbf{E} \big\}$ can be constructed with a set of $K$ Steiner Zone vertices $\mathbf{V} = \big\{ v'_1, ..., v'_K \big\}$ and a corresponding edge set $\mathbf{E} = \big\{ e_{ij}, ... \big\}$, where the edge $e_{ij}$ represents the connection between any two SZ vertices $v'_i$ and $v'_j$. Following \textbf{(CEOP)}, the traveling cost between any two SZ vertices $v'_i$ and $v'_j$ is calculated based on the Euclidean distance cost $\mathscr{C}_d \big( v'_i,\: v'_j \big)$, and the prize for visiting a SZ is $\mathscr{P}(\Omega_i)$. To represent the start and end depot, we introduce two additional points, $v'_0$ and $v'_{K+1}$. Note that $\mathscr{P}(v'_0) = \mathscr{P}(v'_{K+1}) = 0$. The objective of SOP is to find a path starting from the start depot $v'_0$, maximizing collected prizes, and ending at the end depot $v'_{K+1}$, without violating the given distance budget $\mathscr{B}_d$ constraint. Note that in our SOP, each SZ $\Omega_j$ must be visited no more than once, with the visitation necessitating a stop at one of its SZ vertices $v'_i \in \Omega_j$. The formulation of our SOP is provided in \autoref{sec-appendix}.

\subsection{CEOP with non-uniform neighborhoods (CEOP-$\mathcal{N}$)}
Let $\mathbf{G} = \big\{ \mathbf{C},\: \mathbf{E} \big\}$ be a complete graph with a set of $N$ target circles $\mathbf{C} = \big\{ c_1, ..., c_N \big\}$ and a corresponding edge set $\mathbf{E} = \big\{ e_{ij} \big\}$, where the edge $e_{ij}$ represents the connection between target circles $c_i$ and $c_j$. Each target circle $c_i$ is characterized by its center $v_i$, radius $r_i$, and prize $\mathscr{P}(c_i)$. The prize $\mathscr{P}(c_i)$ can be collected at any waypoint $v'$ located within the circular neighborhood (or on its boundary). Similar to \textbf{(CEOP)}, we use $v'_0$ and $v'_{N+1}$ to represent the start and end depot, and $\mathscr{P}(v'_0) = \mathscr{P}(v'_{N+1}) = 0$. The objective of CEOP-$\mathcal{N}$ is to find a path starting from the depot $v'_0$, maximizing collected prizes, and ending at $v'_{N+1}$, without violating the given budget $\mathscr{B}_{f}$ constraint. CEOP-$\mathcal{N}$ introduces two distinct cost functions, $\mathscr{C}_{f_1}\big(v' \:|\: c_j \big)$ for the prize-collecting cost of the target circle $c_j$ at some waypoint $v'$, and $\mathscr{C}_{f_2}\big( v'_i,\: v'_j \big)$ for the traveling cost between waypoints $v'_i$ and $v'_j$. Note that $f_1$ and $f_2$ are two `nominal' functions, implying they can adopt any continuous form. For instance, in the standard CEOP, $\mathscr{C}_{f_1}\big(v' \:|\: c_j \big) = f_1 \big(v',\: c_j \big) = 0$ and $\mathscr{C}_{f_2}\big( v'_i,\: v'_j \big) = f_2\big( v'_i,\: v'_j \big) = \sqrt{(x'_i - x'_j)^2 + (y'_i - y'_j)^2}$. The overall path cost comprises the prize-collecting and traveling costs, constrained by the budget $\mathscr{B}_{f}$. Given decision variables $X_{ij} \in \big\{ 0, 1 \big\}$ (decide the visitation of all target circles and depots), alongside continuous waypoint coordinates and quadratic constraints, we formulate CEOP-$\mathcal{N}$ as a MINLP problem:
\vspace*{-0.1cm}
\begin{subequations}
\begin{flalign}
    \label{eq:ceop-nu-obj}
    \textbf{(CEO}&\textbf{P-}\boldsymbol{\mathcal{N}}\textbf{)}\quad  \max~\sum_{i=0}^N\sum_{j=1}^N \mathscr{P}(c_j) \cdot X_{ij}&\\
    \label{eq:ceop-nu-cost-to-collect-prize}
    \textbf{s.t.}\quad & \mathscr{C}_{f_1}\big(v' \:|\: c_j\big) = f_1\big(v',\: c_j\big)\;, & v' \in c_j,\; j = 1, ..., N&\\
    \label{eq:ceop-nu-cost-to-travel}
    &\mathscr{C}_{f_2} \big(v'_i,\; v'_j\big) = f_2\big(v'_i,\: v'_j\big)\;, & v'_i \in c_i,\; v'_j \in c_j,\; i, j = 1, ..., N,\; i\neq j&\\
    \label{eq:ceop-nu-budget}
    &\sum_{i=0}^{N}\sum_{j=1}^{N} \mathscr{C}_{f_1} \big(v' \:|\: c_j\big) \cdot X_{ij} +  \sum_{i=0}^{N} \sum_{j=1}^{N+1} \mathscr{C}_{f_2} \big(v'_i,\; v'_j\big) \cdot X_{ij} \leq \mathscr{B}_f&\\
    \nonumber
    &\text{Constraint}~\eqref{eq:ceop-start-end},~\eqref{eq:ceop-one-visit},~\eqref{eq:ceop-subtour},~\eqref{eq:ceop-radius}&
\end{flalign}
\end{subequations}
The objective function \eqref{eq:ceop-nu-obj} maximizes the collected prizes. Constraint \eqref{eq:ceop-nu-cost-to-collect-prize} enforces the cost to collect the prize of $c_j$ at waypoint $v'$ must satisfy a continuous function $f_1\big(v',\: c_j\big)$. Constraint \eqref{eq:ceop-nu-cost-to-travel} enforces the cost to travel from waypoint $v'_i$ to waypoint $v'_j$ must satisfy another continuous function $f_2\big(v'_i,\: v'_j\big)$. Constraint \eqref{eq:ceop-nu-budget} implies the overall path cost, formed by costs to collect prizes and costs to travel, must not exceed the given budget $\mathscr{B}_f$. Moreover, there is no prize-collecting cost for the start and end depot.

\subsubsection{Truck-and-drone Delivery Problem (TDDP)} \label{sec:TDDP}
$\textbf{(CEOP-}\boldsymbol{\mathcal{N}}\textbf{)}$ is insufficient when introducing overlapped neighborhoods. As such, we support the above $\textbf{(CEOP-}\boldsymbol{\mathcal{N}}\textbf{)}$ with a straightforward truck-and-drone delivery example. Given the constraint that delivery UAV possesses a maximum operational range denoted as $r_{\text{drone}}$, TDDP delineates the feasible service region (i.e., neighborhood) for each customer $v_i$ as a target circle $c_i$. This circle is characterized by its center at coordinates $(x_i, y_i)$ and a radius equivalent to $r_{\text{drone}}$. We construct a complete graph $\mathbf{G} = \big\{ \mathbf{C},\: \mathbf{E} \big\}$, where $\mathbf{C} = \big\{c_1, ..., c_{N} \big\}$ denotes the set of $N$ target circles and $\mathbf{E} = \big\{ e_{ij}, ... \big\}$ is the corresponding edge set. Each edge $e_{ij}$ represents the connection between target circles $c_i$ and $c_j$. The proximity between customers may result in an overlapping (i.e., SZ) of their respective serviceable areas. Note that the set of Steiner Zoness $\mathbf{Z} = \big\{ \Omega_1, ..., \Omega_M \big\}$ can be obtained by solving \textbf{(SZSP)}. The delivery truck can halt at a designated waypoint $v'$ within such overlap region $\Omega_i$. From this position, it can deploy a maximum of $N_{\max}^{|\Omega|}$ UAVs to concurrently serve proximate customers encompassed by $\Omega_i$. 

The operational protocol for UAV delivery is as follows: Upon mounting the package intended for customer $c_i$, the UAV departs from the truck. The UAV initiates the service phase once arriving above the specified delivery coordinate $(x_i, y_i)$. This phase includes landing, handing over the package, and subsequent takeoff. The duration of this service phase is quantified as a constant $t_{\text{serv}}$. After the service phase, the UAV returns to the truck and prepares for the next delivery. For the customer $v_i$, we use $\mathscr{P}(c_i)$ to denote the delivery priority (i.e., prize). Thus, any waypoint $v' \in \Omega_i$ can collect prize $\mathscr{P}\big(v' \:|\: \Omega_i\big)$ based on Constraint \eqref{eq:sz-prize}. To reflect the heterogeneity of neighborhoods in CEOP-$\mathcal{N}$, TDDP assumes the package weight has a non-trivial impact on the UAV's average flight speed $\mathcal{V}_\text{drone}$. The flight efficiency is quantified as $\lambda_i$ (in percentage). This efficiency parameter ranges between $\big[ \lambda_{\min},\: \lambda_{\max} \big]$, and its magnitude is contingent upon the weight of the package requested by customer $v_i$. As a result, the effective average flight speed of the UAV when delivering the package for customer $v_i$ is calculated as $\lambda_{i}\cdot\mathcal{V}_{\text{drone}}$. Additionally, we denote the average truck speed as $\mathcal{V}_\text{truck}$. Similar to $\textbf{(CEOP-}\boldsymbol{\mathcal{N}}\textbf{)}$, we use $v'_0$ and $v'_{N+1}$ to represent the start and end depot, and $\mathscr{P}(v'_0) = \mathscr{P}(v'_{N+1}) = 0$. TDDP aims to find a path for the truck to start from the depot $v'_0$, maximize collected prizes, and end at depot $v'_{N+1}$, without violating the given time budget $\mathscr{B}_t$ constraint. The truck's traveling and idle time defines a path's overall time cost $\mathscr{C}_t$. With decision variables $X_{ij} \in \big\{ 0, 1 \big\}$ that determine the visitation of all SZs and depots, the TDDP can be formulated as follows:
\begin{subequations}
\begin{flalign}
    \label{eq:tddp-obj}
    \textbf{(TDDP)}\quad & \max~\sum_{i=0}^N \sum_{j=1}^{N} \mathscr{P}\big( v' \:|\: \Omega_j \big) \cdot X_{ij} &\\
    \label{eq:tddp-sz-max-degree}
    \textbf{s.t.}\quad & |\Omega_i| \leq N_{\max}^{|\Omega|}\: , & i=1,..., M &\\
    \label{eq:tddp-cost-to-collect-prize}
    & \mathscr{C}_t\big( v' \:|\: \Omega_j \big) = \max\bigg\{ \frac{\mathscr{C}_d(v',\: v_i)}{\lambda_{i}\cdot\mathcal{V}_{\text{drone}}} + t_\text{serv} \:+ \frac{\mathscr{C}_d(v_i,\: v')}{\mathcal{V}_{\text{drone}}} \: : \forall\: v_i \in \Omega_j \bigg\}\: , & j=1,..., M &\\
    \label{eq:tddp-cost-to-travel}
    & \mathscr{C}_t \big( v'_i,\; v'_j \big) = \frac{\mathscr{C}_d\big( v'_i,\: v'_j \big)}{\mathcal{V}_\text{truck}}\;, \;& i, j = 1, ..., M,\; i\neq j&\\
    \label{eq:tddp-prize}
    & \mathscr{P}\big( v' \:|\: \Omega_{j} \big) = \sum_{c_i\in\:\Omega_j} \mathscr{P}(c_i)\; , & j = 1, ..., M &\\
    \label{eq:tddp-budget}
    & \sum_{i=0}^{N}\sum_{j=1}^{N} \mathscr{C}_t\big( v' \:|\: \Omega_j \big) \cdot X_{ij} + \sum_{i=0}^{N} \sum_{j=1}^{N+1} \mathscr{C}_t\big( v'_i,\; v'_j \big) \cdot X_{ij}\:\leq\:\mathscr{B}_t &\\
    \nonumber
    &\text{Constraints}~\eqref{eq:ceop-start-end},~\eqref{eq:ceop-one-visit},~\eqref{eq:ceop-subtour},~\eqref{eq:sz-constraint}&
\end{flalign}
\end{subequations}
The objective function \eqref{eq:tddp-obj} maximizes the prize of all packages delivered in the path. Constraint \eqref{eq:tddp-sz-max-degree} enforces the maximum degree of any SZ cannot exceed the number of UAVs $N_{\max}^{|\Omega|}$. Constraint \eqref{eq:tddp-cost-to-collect-prize} defines the time cost of the truck waiting at the waypoint as the maximum time among all UAV deliveries, where $t_\text{serv}$ is a constant representing the general UAV service time. Constraint \eqref{eq:tddp-cost-to-travel} defines the time cost for the truck to travel from waypoint $v'_i$ to $v'_j$, where $v'_i$ and $v'_j$ must be located in different SZs. Constraint \eqref{eq:tddp-prize} defines a SZ $\Omega_k$'s prize $\mathscr{P}(v' \: | \: \Omega_k)$, which can be collected at a waypoint $v'$, as the prize sum of all target customers included in the SZ. Constraint \eqref{eq:tddp-budget} enforces the overall time cost of the path, formed by a truck's idle and traveling time, must not exceed the given time budget $\mathscr{B}_t$. Moreover, there is no prize-collecting cost for the start and end depot.

\section{System Design} \label{sec:system-design}
\begin{figure}[!b]
\centering
\begin{minipage}[t]{0.35\linewidth}
    \centering
    \includegraphics[width=\textwidth]{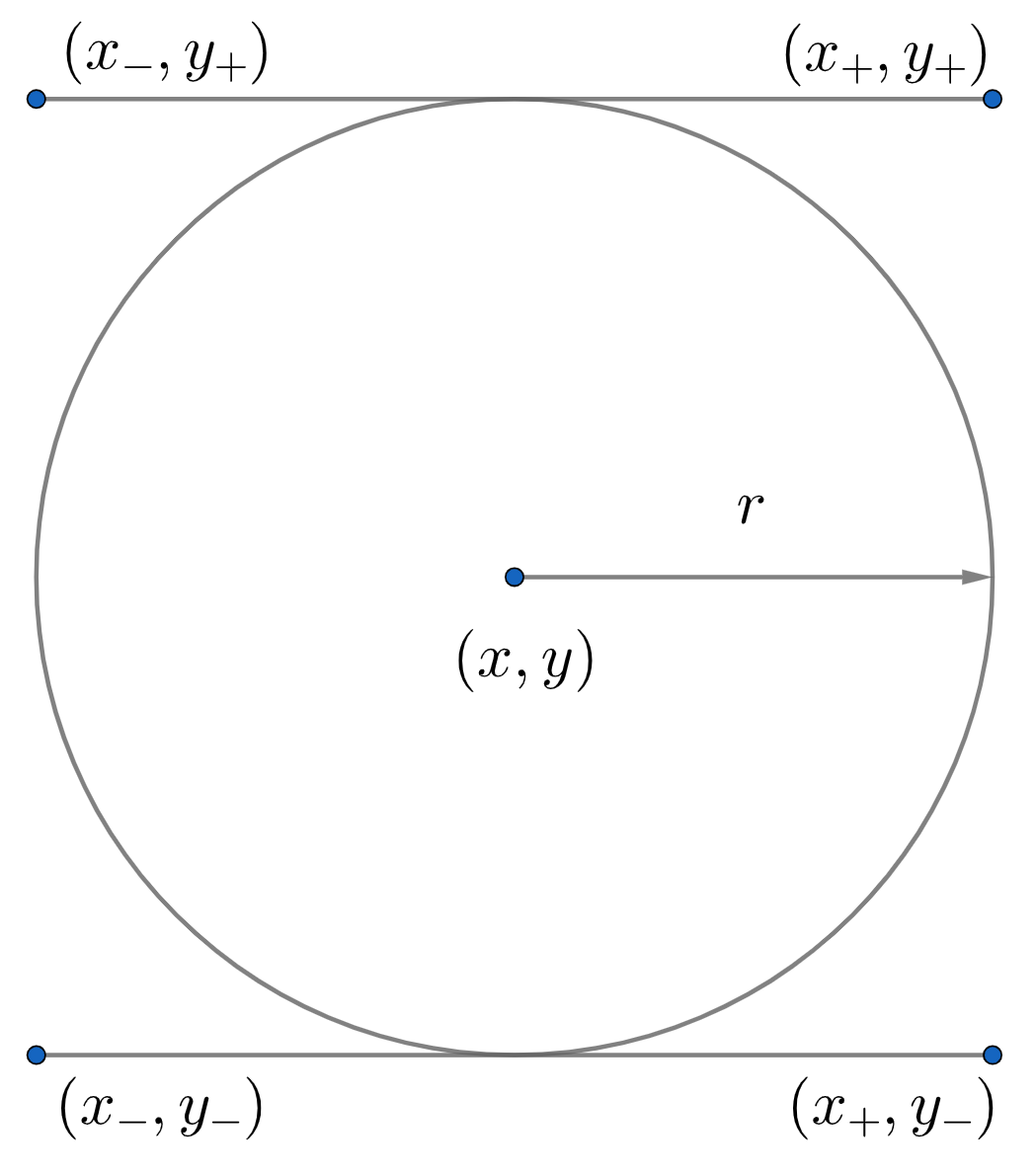}
    \subcaption{Circle approximation~\citep{wang2019steiner}} \label{fig:wang-circle-approx}
\end{minipage}
\hspace{0.5cm}
\begin{minipage}[t]{0.55\linewidth} 
    \includegraphics[width=\textwidth]{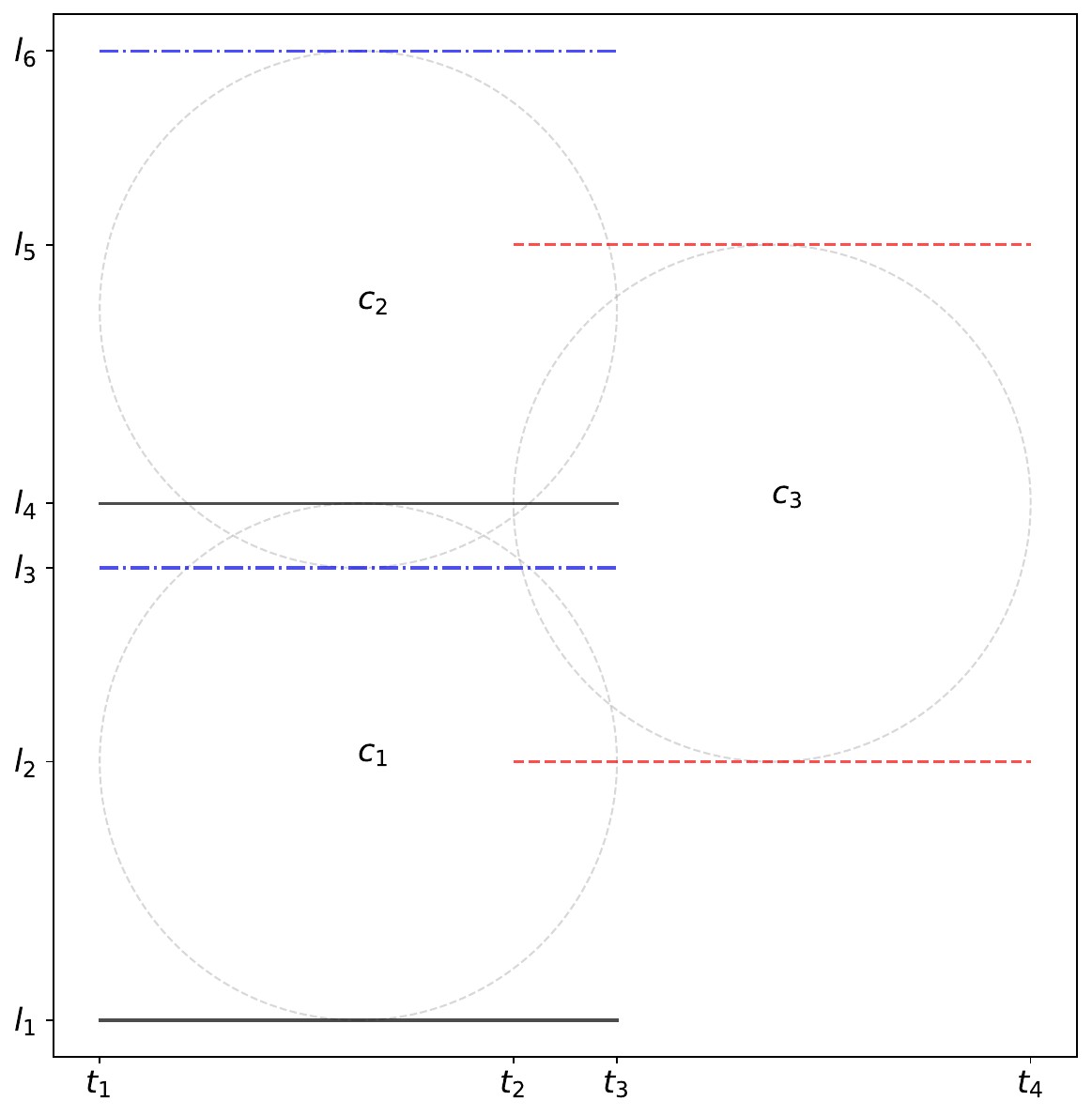}
    \subcaption{An example that the sweep line algorithm will find a wrong SZ of degree three. The interval $[l_2, l_5]$ is registered at $t_2$, overlapping with $[l_3, l_6], [l_1, l_4]$ and $[l_3, l_4]$. Three circles intersect pairwise. But in fact, only a SZ of degree two can be formed.} \label{fig:nece_no_suff}
\end{minipage}
\caption{Sweep line algorithm~\citep{wang2019steiner} and its limitation}
\end{figure}
This section explains the design of CRaSZe-AntS for CEOP and CEOP-$\mathcal{N}$, respectively. Following the framework stated in \S \ref{sec:related-work}, the first step is discretizing targets with circular neighborhoods. CRaSZe-AntS employs the RSZD scheme to construct a valid SZ layout and record all SZ vertices as candidates. CRaSZe-AntS is adaptable for both uniform and non-uniform neighborhood scenarios within CEOP and CEOP-$\mathcal{N}$. This versatility is enabled through two distinct modes:
\begin{itemize}
    \item In CEOP, searching the interior of constructed SZs is not compulsory; CRaSZe-AntS, therefore, utilizes ACS (with additional add/drop operators) to yield a solution from the resultant SOP instance. Finally, the arc search algorithm and the add operator optimize the discrete solution using the geometry of SZs.
    \item In CEOP-$\mathcal{N}$, only continuous searching of SZs' interiors can evaluate the minimum cost taken to collect prizes, thereby formulating an OP scenario instead. CRaSZe-AntS leverages PSO to refine waypoint positions based on the constructed SZ layout. IACS (with add/drop operators), served as the solver for OP instances, are embedded within all particles of the PSO. This dual-layered strategy facilitates the iterative refinement of solutions, where the outcomes from IACS guide the update processes for both the \textit{positions} and \textit{velocities} of all particles.
\end{itemize}
\begin{algorithm}[!t]
    \SetKwInOut{Input}{Input}
    \SetKwInOut{Output}{Output}
    \caption{Randomized Steiner Zone Discretization scheme} \label{alg:rszd}
    \Input{Circle set $\mathbf{C} = \big\{ c_1, ..., c_N \big\}$; Number of iterations $N_{\text{iter}}^{\text{RSZD}}$; Maximum number of circles $N_{\max}^{|\Omega|}.$}
    \For{$t = 1$ to $N_{\mathrm{iter}}^{\mathrm{RSZD}}$}{
        Randomly shuffle $\mathbf{C}$ if not the first iteration, i.e., $t \neq 1$\;
        Initialize the feasible circle set $\mathbf{A}_\mathbf{C} \leftarrow \mathbf{C}$\;
        \For{each circle $c_i \in \mathbf{A}_\mathbf{C}$}{
            Create a new SZ $\Omega_m \leftarrow \{ c_i \}$\;
            Remove $c_i$ from feasible circle set $\mathbf{A}_\mathbf{C} \leftarrow \mathbf{A}_\mathbf{C} \setminus \{ c_i \}$\;
            \For{each remaining circle $c_j \in \mathbf{A}_\mathbf{C}$}{
                \If{$|\Omega_m| \:<\: N^{\Omega}_{\max}$}{
                    $\forall c' \in \Omega_m$, if distance $\mathscr{C}_d(c', c_j) < r_{c'} + r_{c_j}$, then $c_j$ passes the necessity condition. Otherwise, skip to the next circle in $\mathbf{A}_\mathbf{C}$\;
                    $\forall c' \in \Omega_m$, if there exists at least one intersection point (from $c'$ and $c_j$) located in all other circles in the SZ $\Omega_m$, then $c_j$ passes the sufficiency condition\;
                    \If{$c_j$ passes the sufficiency condition}{ 
                        Update the corresponding vertex list of the SZ $\Omega_m$\;
                    }
                    \If{$c_j$ passes \textbf{all} conditions}{
                        $\Omega_m \leftarrow \Omega_m \cap \{ c_j \}$\;
                        $\mathbf{A}_\mathbf{C} \leftarrow \mathbf{A}_\mathbf{C} \setminus \{ c_j \}$\;
                    }
                }
            }
        }
        Record current SZ set $\mathbf{Z}_t$\;
    }
    \Return The SZ set $\mathbf{Z}_i = \big\{ \Omega_1, ..., \Omega_M \big\}$, where $i = \arg\min |\mathbf{Z}_t|$, $t = 1, ..., N_{\text{iter}}^{\text{RSZD}}$, and $M = |\mathbf{Z}_i|$.
\end{algorithm}

\subsection{Randomized Steiner Zone Discretization (RSZD) scheme}
\citet{wang2019steiner} proposed the sweep line algorithm for recognizing all possible combinations of SZs. The final layout of SZs is determined by solving the set covering problem. As shown in Fig. \ref{fig:wang-circle-approx}, the circle is approximated to two line segments $\overline{(x_{-},y_{+})(x_{+},y_{+})}$ and $\overline{(x_{-}, y_{-})(x_{+},y_{-})}$. Each circle's $x_{-}$ and $x_{+}$ are recorded as $t_1, t_2, ...$ (from left to right); each circle's $y_{-}$ and $y_{+}$ are recorded as $l_1, l_2, ...$ (from bottom to top). The algorithm employs a vertical line that sweeps from $t_1$ to check any potential overlap in each interval (e.g., $\big[t_1, t_2\big]$, $\big[t_2, t_3\big]$, etc.) and register associated SZs. However, the sweep line algorithm exhibits three significant limitations, specifically, it: 
\begin{enumerate}
    \item can be computationally demanding in scenarios featuring high-degree SZs \citep[][Appendix A.]{wang2019steiner}.
    \item cannot locate the vertices of SZs.
    \item does not provide an explanation for addressing some extreme scenarios (e.g., as shown in Fig.~\ref{fig:nece_no_suff}). A wrong SZ of degree three would be identified because $\big[l_1, l_4\big], \big[l_3, l_6\big], \big[l_2, l_6\big]$ are registered at the interval $\big[t_2, t_3\big]$, and $c_1, c_2, c_3$ intersect with each other.
\end{enumerate}

\begin{figure*}[!t]
    \centering
    \includegraphics[width=\textwidth]{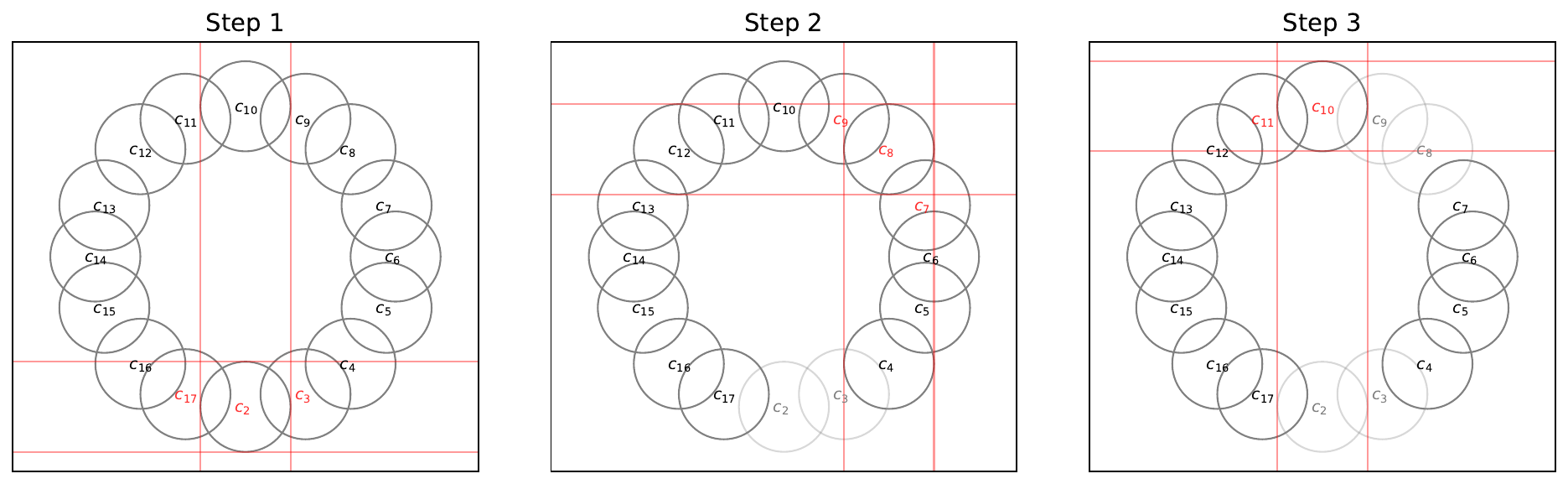}
	\caption{An example of the first three steps of RSZD in \textit{concentricCircles1}.}
	\label{fig:rszd}
\end{figure*}

We introduce the Randomized Steiner Zone Discretization (RSZD) scheme (Alg. \ref{alg:rszd}) to overcome the aforementioned limitations. In contrast to the sweep line algorithm, RSZD randomly selects a candidate circle to establish a SZ. It then evaluates its neighboring circles to determine whether they can be incorporated into this SZ based on \textbf{Condition} \ref{prop:suff} and \ref{prop:nece}. Fig. \ref{fig:rszd} gives an example of the first three steps of RSZD in constructing SZs for \textit{concentricCircles1} \citep{mennell2009heuristics}. 

\subsection{Approach to CEOP with uniform neighborhoods} \label{sec:ceop-acs-arc}
TSP and its variants are frequently addressed using two discrete metaheuristics: ACS \citep{dorigo1997ant} and GA \citep{mitchell1998introduction} due to their established presence in literature, straightforward implementation, and high adaptability and effectiveness for various instances of TSP, OP, and related extensions \citep{silva2008rescheduling, mandal2020survey}. As presented by \citet{liang2006ant}, the ant colony optimization algorithm can yield a solution with comparable quality within a significantly reduced time compared to GA in the OP scenario. Furthermore, GA's adaptation to the OP scenario typically necessitates additional adjustments to its crossover operator because the solution length is variable during evolution. Therefore, once RSZD discretized CEOP, we adapted ACS to tackle the resultant SOP. Importantly, RSZD is compatible with a broad range of existing methods for the SOP (i.e., not limited to ACS). This flexibility stems from RSZD's output, specifically, a set of SZs and their vertices that naturally form a SOP scenario. ACS simulates the foraging behavior of an ant colony, incorporating three fundamental rules: the state transition rule, which decides the next visitation; the local updating rule, responsible for adjusting the pheromone trail visited by all ants; and the global updating rule, which updates the pheromone matrix based on the global-best ant. 

In our state transition rule, the probability for the ant $k$ at the waypoint $v'_r \in \Omega_i$ to visit the next waypoint $v'_s \in \Omega_j$ is defined as:
\begin{equation} \label{eq:acs-random-prob}
    p_k(r, s) = \begin{dcases}
                \dfrac{\big[ \tau(r,\: s) \big] \cdot \big[ \eta(r,\: s) \big]^{\beta}}{\sum_{v'_u \in\: \mathbf{A}_k^{\Omega}}\: \big[ \tau(r,\: u) \big] \cdot \big[ \eta(r,\: u) \big]^{\beta}}\;, & \text{if}~v'_s \in \mathbf{A}_k^{\Omega}\\
                0\;, & \text{otherwise}
            \end{dcases}
\end{equation}
where $\tau(r, s)$ is the pheromone deposited on edge $e_{rs}$. We define the heuristic information $\eta(r, s) = \dfrac{\mathscr{P}(v'_s \:|\: \Omega_j)}{ \mathscr{C}_d(v'_r,\: v'_s)}$ as the ratio of the SZ $\Omega_j$'s prize to the distance cost between these two waypoints. $\beta$ is a parameter to control the relative importance of pheromone versus heuristic information. We denote the feasible set of remaining SZs in the ant $k$ by $\mathbf{A}_k^{\Omega}$. Note that our SOP allows at most one visitation to the same SZ. To balance exploring and exploiting, the state transition rule introduces an additional parameter $q_0 \in \big( 0, 1 \big)$: 
\begin{equation} \label{eq:acs-state}
    s = \begin{dcases}
        \arg\max_{v'_s \in\: \mathbf{A}_k^{\Omega}} \Big\{ \big[ \tau (r,\: s) \big] \cdot \big[ \eta(r, \: s) \big]^{\beta} \Big\}\; , & q \leq q_0\\
        s \sim p_k(r,\: s) ~\text{in Eq.~\eqref{eq:acs-random-prob}} \; , & q > q_0
    \end{dcases}
\end{equation}
The probability $q \in \mathbb R$ is randomly generated from a uniform distribution ranging in $[0, 1]$.

To reduce the probability of ants constructing the same solution, the local updating rule is applied to edges visited by ants after the solution construction phase:
\begin{eqnarray}\label{eq:acs-local}
    \tau(r, s) \leftarrow (1-\rho)\cdot \tau(r, s) + \rho \cdot \tau_0(r, s)
\end{eqnarray}
The evaporation rate $\rho \in (0, 1)$ is a constant that limits the accumulated pheromone on edge $e_{rs}$. \citet{liang2006ant} suggested to set the initial pheromone as $\tau_0 = (N \cdot \mathscr{B}_d)^{-1}$, where $N$ is the number of all target nodes, while keeping heuristic information as $\eta = \mathscr{P} / \mathscr{C}_d$. Such a setting may cause significant variations in pheromone levels at the beginning of ACS's evolution, potentially deterring ants from exploring the search space. Following the setting by \citet{dorigo1997ant}, we employ the nearest neighbor heuristic to generate a feasible initial solution. To remain consistent with the definition of heuristic information, the initial pheromone is set to $\tau_0 = \mathscr{P} / (N_{wp} \cdot \mathscr{C}_d)$, where $N_{wp}$ is the number of waypoints in the path, $\mathscr{P}$ and $\mathscr{C}_d$ are the path prize and cost, respectively. Here, $\tau_0$ is divided by $N_{wp}$ instead of $N$ because the solution length in SOP is variable, while in TSP, the solution length always equals to $N$. 

In ACS, only the global-best ant, whose solution achieves the highest quality so far (i.e., either maximum prizes or minimum costs when prizes are the same), can deposit the pheromone at the end of each iteration. The global updating rule is defined as: 
\begin{eqnarray}\label{eq:acs-global}
    \tau(r,\: s) \leftarrow (1-\alpha)\cdot \tau(r,\: s) + \alpha \cdot \Delta \tau(r,\: s)
\end{eqnarray}
where $\alpha \in (0, 1)$ is a constant to control the pheromone decay rate, the deposited pheromone can be obtained by:
\begin{eqnarray} \label{eq:acs-delta-tau-ceop}
    \Delta\tau(r, s) = \begin{dcases} \mathscr{P}^{gb} \: / \: \mathscr{C}_d^{gb} \; , & \text{if $e_{rs} \in $ global-best path}\\
        0 \; , & \text{otherwise}
    \end{dcases}
\end{eqnarray}
$\mathscr{P}^{gb}$ and $\mathscr{C}_d^{gb}$ are the collected prize and traveling cost of the global-best path, respectively. We opted for a straightforward 2-opt local search method for later path sequence improvement. The pseudo-code block of ACS to solve SOP can be seen as Alg.~\ref{alg:acs-sop}.

\begin{algorithm}[!b]
    \caption{Drop operator} \label{alg:drop-operator}
    \SetKwInOut{Input}{Input}%
    \Input{Ant path $\{ v'_1, ..., v'_{N_{wp}}\}$;\: Feasible node set $\mathbf{A}^{v'}$.}
    \While{ant path does \textbf{not} satisfy Constraint \eqref{eq:ceop-budget}}{
        \For{each waypoint $v'$ in ant path (exclude start and end waypoint)}{
            Compute drop value $drop(v')$ by Eq. \eqref{eq:drop-value}\;
        }
        Find the path index $i$ at which the node has the minimum drop value, i.e., $i = \arg\min \big\{ drop (v'_j), ... \big\}$\;
        Update path cost $\mathscr{C}_d \leftarrow \mathscr{C}_d - \mathscr{C}_d(v'_{i-1},\: v'_i) - \mathscr{C}_d(v'_i,\: v'_{i+1}) + \mathscr{C}_d(v'_{i-1},\: v'_{i+1})$\;
        Update path prize $\mathscr{P} \leftarrow \mathscr{P} - \mathscr{P}(v'_{i} \: |\: \Omega_k)$\;
        Remove the node at path index $i$\;
        Update the feasible set $\mathbf{A}^{v'} \leftarrow \mathbf{A}^{v'} \cup \big\{ v_i' \:|\: \forall v_i' \in \Omega_k \big\}$\;
    }
    \Return The feasible path with the new prize, new cost, and updated feasible node set.
\end{algorithm}

\subsubsection{Local search operators} \label{sec:local-search}
\begin{algorithm}[!b]
    \caption{Add operator} \label{alg:add-operator}
    \SetKwInOut{Input}{Input}%
    \Input{ant path $\{ v'_1, ..., v'_{N_{wp}}\}$;\: Feasible node set $\mathbf{A}^{v'}$.} 
    \While{exist any node $\in \mathbf{A}^{v'}$ can be inserted into ant path without violating Constraint \eqref{eq:ceop-budget}}{
        \For{each node $v'_k \in \mathbf{A}^{v'}$}{
            Get 3 neighbor nodes in the ant path with minimum distance cost to visit $\mathbf{S}_{\text{nbr}} =\big\{ nbr_1,\: nbr_2,\: nbr_3 \big\}$\;
            \For{each pair of neighbor nodes $(nbr_i,\: nbr_j) \in \mathbf{S}_{\mathrm{nbr}}$}{
                Check whether this pair is adjacent in the ant path\;
            }
            \If{\textbf{no} adjacent pair exists}{
                \For{$nbr_i \in \mathbf{S}_{\mathrm{nbr}}$}{
                    Find the previous and next node of $nbr_i$ in the path\;
                    Create new pair $(v'_{prev},\: nbr_i)$ and $(nbr_i,\: v'_{next})$\;
                }
            }
            Find the pair that minimizes the visitation cost and find at which index $l$ to insert\;
            Compute the add value $add(v'_k \:|\: l)$ by Eq. \eqref{eq:add-value}\;
        }
        Find the waypoint $v'_j$ with maximum add value and its insert index $i$ in the path, i.e., $(j, i) = \arg\max \big\{ add(v'_k \:|\: l), ... \big\}$\;
        Update path cost $\mathscr{C}_d \leftarrow \mathscr{C}_d + \mathscr{C}_d(v'_{i-1},\: v'_j) + \mathscr{C}_d(v'_j, v'_{i+1}) - \mathscr{C}_d(v'_{i-1},\: v'_{i+1})$\;
        Update path prize $\mathscr{P} \leftarrow \mathscr{P} + \mathscr{P}(v'_j \:|\: \Omega_m)$\;
        Insert the node $v'_j$ into the ant path (at path index $i$)\;
        Update the feasible set $\mathbf{A}^{v'} \leftarrow \mathbf{A}^{v'} \setminus \big\{ v_j' \:|\: \forall v_j' \in \Omega_m \big\}$\;
    }
    \Return The feasible path and the updated feasible node set.
\end{algorithm}
\begin{figure*}[!t]
    \centering
    \begin{subfigure}{0.3\textwidth}
        \includegraphics[width=\textwidth]{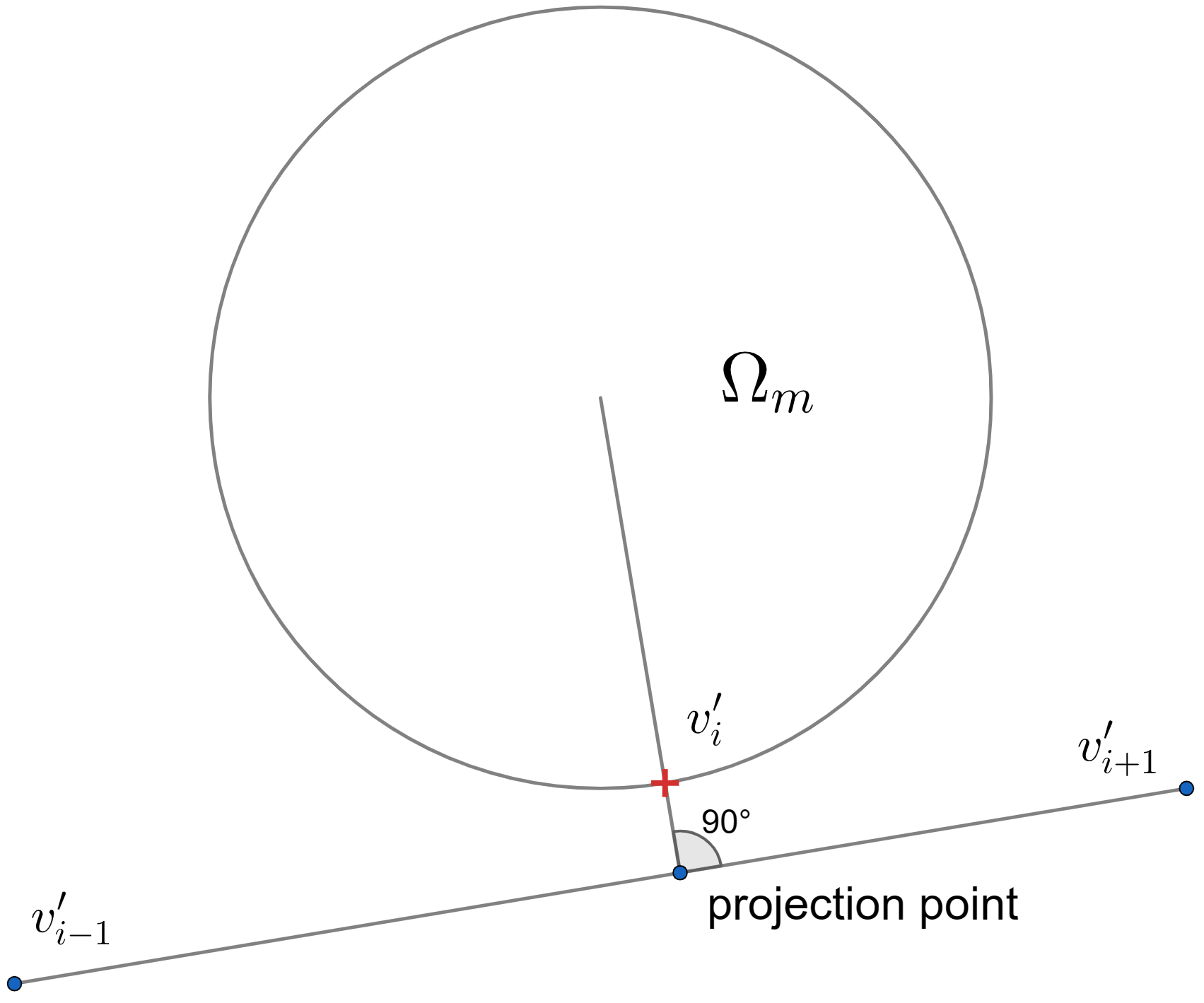}
	\subcaption{When the line segment has no intersection with the SZ of degree 1.}
	\label{fig:arc-search-degree1}
    \end{subfigure}
    \hspace*{0.1cm}
    \begin{subfigure}{0.33\textwidth}
        \includegraphics[width=\textwidth]{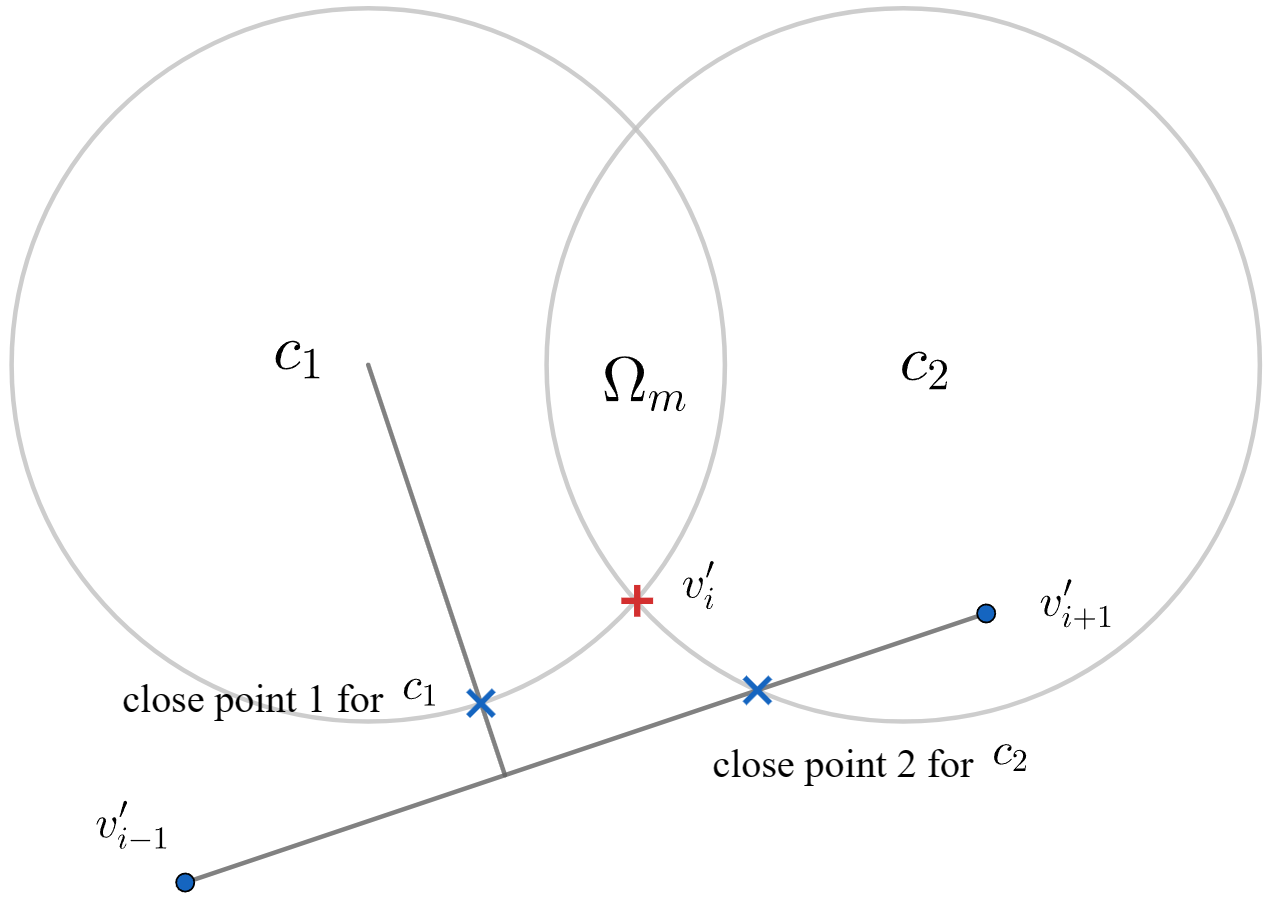}
        \subcaption{When the line segment has no intersection with the SZ of degree 2.}
    \end{subfigure}
    \hspace*{0.1cm}
    \begin{subfigure}{0.33\textwidth}
        \includegraphics[width=\textwidth]{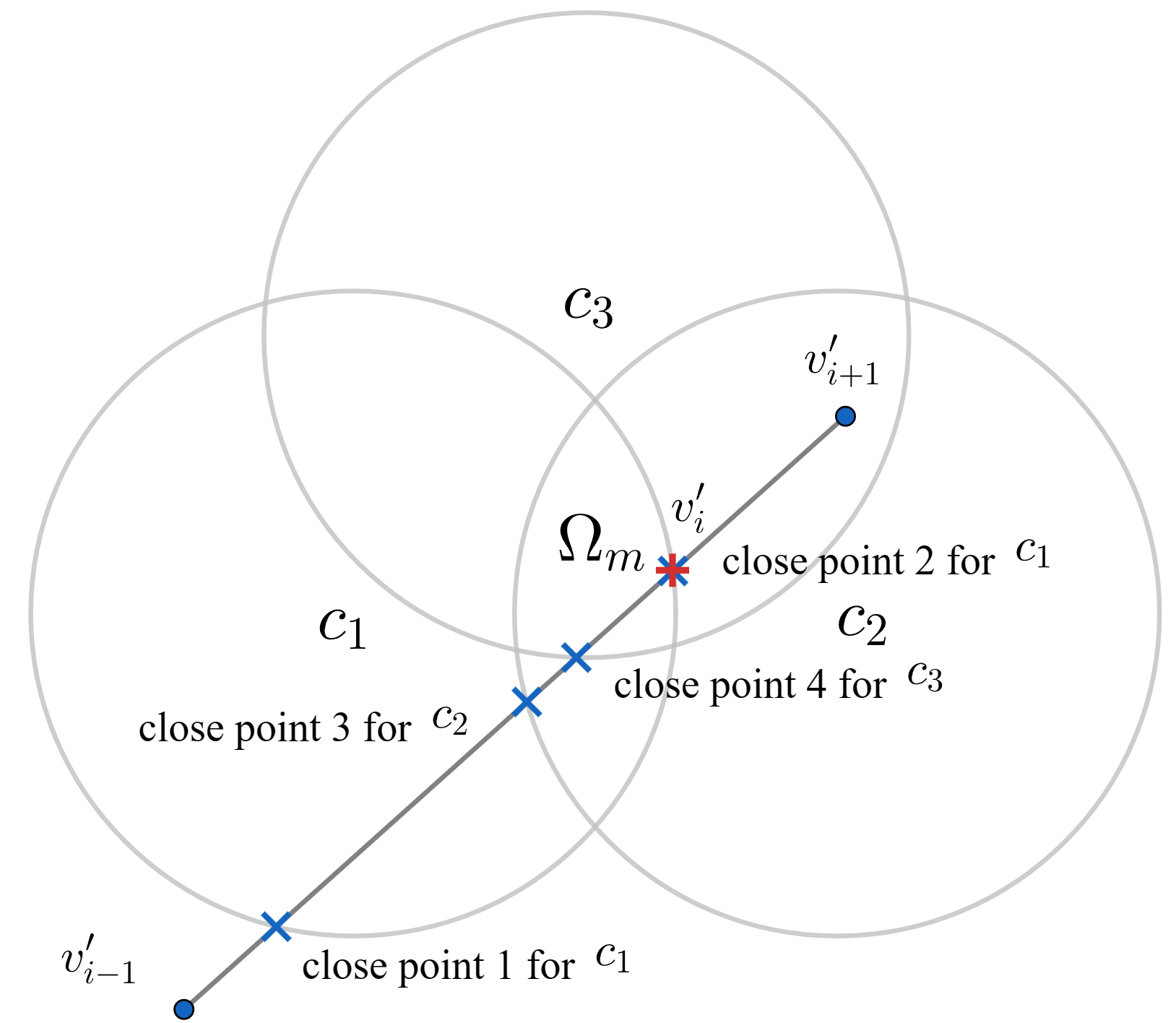}
        \subcaption{When the line segment has two intersection points with the SZ of degree 3.}
    \end{subfigure}
    \caption{Examples of the arc search algorithm for SZ of different degrees.}
\end{figure*}
Additional local search procedures are incorporated to enhance the solution quality. To avoid violating Constraint \eqref{eq:ceop-budget}, we adapted the drop operator from \citep{kobeaga2018efficient} that removes the node with the lowest benefit for visitation (Alg. \ref{alg:drop-operator}). To assess the desirability of dropping a waypoint $v'_i \in \Omega_j$ (if removed at path index $i$), a metric, known as \textit{drop value}, is defined as:
\begin{eqnarray} \label{eq:drop-value}
    drop(v'_i) = \dfrac{\mathscr{P}(v'_i \:|\: \Omega_j)}{\mathscr{C}_d(v'_{i-1},\: v'_i) + \mathscr{C}_d(v'_i,\: v'_{i+1}) - \mathscr{C}_d(v'_{i-1},\: v'_{i+1})}
\end{eqnarray}
The drop operator then eliminates the waypoint associated with the smallest drop value. 

The add operator (Alg. \ref{alg:add-operator}) is sequentially invoked after the drop operator. The add operator selects and inserts one waypoint with the largest \textit{add value}. The add value of a waypoint $v'_j \in \Omega_k$ (if inserted at path index $i$) is defined as:
\begin{eqnarray} \label{eq:add-value}
    add(v'_j \:|\: i) = \dfrac{\mathscr{P}(v'_j \:|\: \Omega_k)}{\mathscr{C}_d(v'_{i-1},\: v'_j) + \mathscr{C}_d(v'_{j},\: v'_{i + 1}) - \mathscr{C}_d(v'_{i-1},\: v'_{i+1})}
\end{eqnarray}

\begin{figure}[!t]
    \centering
    \includegraphics[width=\textwidth]{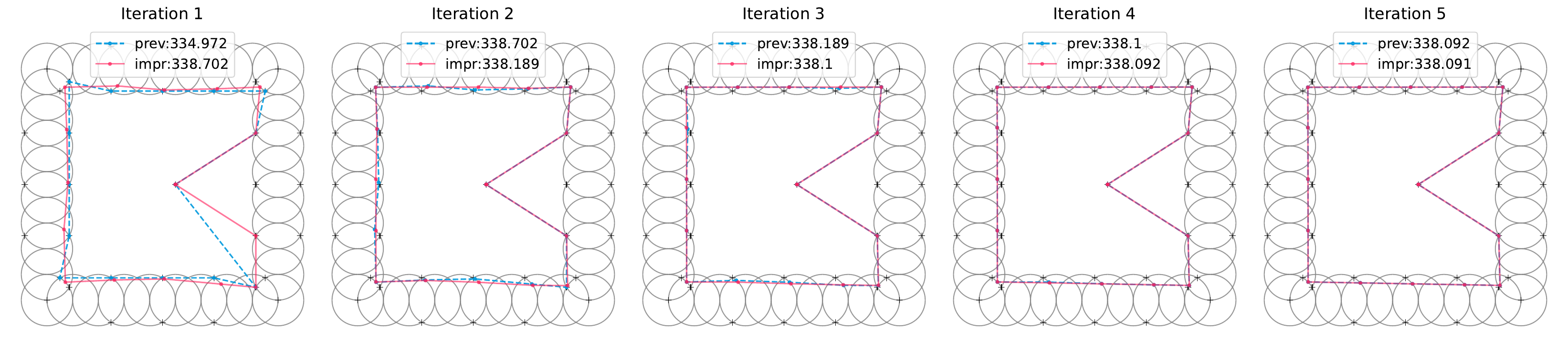}
    \caption{An example of invoking arc search algorithm and add operator for five times on \textit{bubbles1} ($\mathscr{B}_d = 340$). At iteration 1, the arc search algorithm reduced the path cost. This allows the add operator to insert an additional waypoint.} \label{fig:arc-search}
\end{figure}

Upon determining the path sequence, we employ an efficient local search operator, named arc search algorithm (Alg. \ref{alg:arc-search}), to optimize the position of continuous waypoint $v'_i \in \Omega_j$ according to their previous waypoint $v'_{i-1}$ and next waypoint $v'_{i+1}$ along the path. For single-circle scenarios, i.e., $|\Omega_j| = 1$, there are three relationships between the line segment $\overline{ v'_{i-1} v'_{i+1} }$ and the circle: (1) no intersection points, (2) one intersection point (tangent or intersecting without passing through), and (3) two intersection points. In relationship (1), the optimal waypoint position for $v'_{i}$ must coincide with this intersection point. If $\overline{ v'_{i-1} v'_{i+1} }$ fails to intersect the circle, the arc search algorithm projects the circle center onto $\overline{ v'_{i-1} v'_{i+1} }$, designating the intersection point on the circle as the waypoint position (e.g., as shown in Fig. \ref{fig:arc-search-degree1}). For high-degree SZs, the arc search algorithm first inspects the relationship between $\overline{ v'_{i-1} v'_{i+1} }$ and each circle within the SZ. Subsequently, it identifies \textit{close points} at which the line segment and circle $c_k$. We define close points as:
\begin{eqnarray} \label{eq:close-point-def}
    \text{close points} = \begin{dcases}
        \text{the nearest point to } \overline{ v'_{i-1} v'_{i+1} }\; , & \text{if } \overline{ v'_{i-1} v'_{i+1} } \text{ has no intersection with } c_k\\
        \text{the intersection points on } c_k\; , & \text{ otherwise}
    \end{dcases}
\end{eqnarray}
For each circle, if any close point falls within the feasible range of the circular arc formed by two SZ vertices, this close point is designated as the final position for $v'_i$. Otherwise, the close point is infeasible. Arc search algorithm then selects the nearest SZ vertex as the candidate for that particular circle. Through traversing all circles, the arc search algorithm can determine the optimal position for $v'_i$ among chosen candidates. Subsequently, the add operator is iteratively invoked until the solution attains the maximal budget utilization. Fig. \ref{fig:arc-search} demonstrates how the arc search algorithm and the add operator improve the solution from ACS on \textit{bubbles1}. At iteration 1 (the leftmost sub-figure), the arc search algorithm first relocated waypoints (previously were SZ vertices), and then the add operator inserted a waypoint from the waiting list.
\begin{algorithm}[!t]
    \caption{ACS to solve SOP} \label{alg:acs-sop}
    \SetKwInOut{Input}{Input}%
    \Input{SZ set $\mathbf{Z} = \big\{ \Omega_1, ..., \Omega_M \big\}$;\:Number of ants $N_{\text{ant}}$;\: Number of iterations $N_{\text{iter}}^{\text{ACS}}$;\\
    Maximum number of no improvement $N_{\text{impr}}^{\text{ACS}}$;\: Improvement tolerance $\varepsilon^{\text{ACS}}$;\\ $\beta$ in Eq. \eqref{eq:acs-random-prob};\: $q_0$ in Eq. \eqref{eq:acs-state};\: $\rho$ in Eq. \eqref{eq:acs-local};\: $\alpha$ in Eq. \eqref{eq:acs-global}.}
    Compute initial pheromone $\tau_0$ by nearest neighbor heuristic\;
    Initialize $N_{\text{ant}}$ ants and feasible node set $A^{v'}$, and set no improvement counter to $0$\;
    \For{$n_{it}=1$ to $N_{\mathrm{iter}}^{\mathrm{ACS}}$}{
        \If{no improvement counter $\geq N_{\mathrm{impr}}^{\mathrm{ACS}}$}{ Break the loop\; }
        \For{each ant $k$}{
            Randomly sample the first node to visit $v' \in \mathbf{A}^{v'}$\;
            \While{$\mathbf{A}^{v'} \neq \varnothing$ \textbf{and} ant path satisfies Constraint \eqref{eq:ceop-budget}}{
                Select the next node to visit by Eq. \eqref{eq:acs-state} and add it to the ant path\;
                Update the prize, cost, and feasibility of the path\;
            }
            Add the end waypoint, then update path cost and feasibility\;
            Apply 2-opt operator, then update path cost and feasibility\;
            \If{ant path \textbf{not} feasible}{
                Invoke the drop operator (Alg. \ref{alg:drop-operator})\;
            }
            Invoke the add operator (Alg. \ref{alg:add-operator})\;
            Update the pheromone matrix by Eq. \eqref{eq:acs-local}\;
        }
        Update the local-best ant with index equals to $\arg\max_k \big\{ \mathscr{P}^k \big\}$ (or $\arg\min_k \big\{ \mathscr{C}_d^k\big\}$ if $\mathscr{P}$ is maximum)\;
        \If{\big( $\mathscr{P}^{lb} \geq \mathscr{P}^{gb} + \varepsilon^{\text{ACS}}$ \big) \textbf{or} \big($\mathscr{P}^{lb} = \mathscr{P}^{gb}$ \textbf{and } $\mathscr{C}_d^{lb} \leq \mathscr{C}_d^{gb} - \varepsilon^{\text{ACS}}$ \big)}{
            Update the global-best ant\;
            Update the pheromone matrix by Eq. \eqref{eq:acs-global}\;
        }
        \Else{ No improvement counter $+1$\; }
        Reset all ants (except the global-best ant)\;
    }
    \Return The path sequence, path cost, and path prize of the global-best ant.
\end{algorithm}
\begin{algorithm}[!t]
    \caption{Arc search algorithm} \label{alg:arc-search}
    \SetKwInOut{Input}{Input}%
    \Input{SZ set $\mathbf{Z} = \big\{ \Omega_1, ..., \Omega_M \big\}$; Waypoint sequence from ACS $\big\{ v'_0, ..., v'_{N+1} \big\}$; Number of iterations $N_{\text{iter}}^{\text{arc}}.$}
    \For{$i = 1$ to $N_{\mathrm{iter}}^{\mathrm{arc}}$}{
        \For{$j = 1$ to $N_{wp} - 1$}{
            Record the previous waypoint $v'_{j-1}$, next waypoint $v'_{j+1}$, and the SZ such that $v'_j \in \Omega_k$\;
            \For{each circle $c_l \in \Omega_k$}{
                Compute the `close points' according to the line segment $\overline{v'_{j-1} v'_{j+1}}$ and $c_l$ by Eq. \eqref{eq:close-point-def}\; 
                Compute the feasible arc range for $c_l \in \Omega_k$\;
                \If{any `close point' located in the feasible range}{
                    $v'_j \leftarrow$ the feasible `close point'\;
                }
            }
            \If{\textbf{no} feasible `close point' found}{
                \For{each vertex $v'$ of \:$\Omega_k$}{
                    Calculate the distance $\mathscr{C}_d(v'_{j-1},\: v') + \mathscr{C}_d(v',\: v'_{j+1})$\;
                }
                $v'_j \leftarrow$ the closest vertex\;
            }
            Update the path distance cost\;
        }
    }
    \Return The updated waypoint sequence and the updated path cost.
\end{algorithm}

\subsection{Approach to CEOP-$\mathcal{N}$ with non-uniform neighborhoods}
In CEOP-$\mathcal{N}$, the optimization algorithm must explore the interior regions of overlapped neighborhoods. Our approach to TDDP employs a combination of techniques: RSZD for SZ construction, PSO for waypoint positioning, and IACS for optimizing the sequence of waypoints (by solving the OP). The pseudo-code block of applying PSO and IACS to solve TDDP is shown as Alg. \ref{alg:PSO-IACS-TDDP}.
\begin{algorithm}[!b]
    \caption{PSO and IACS to solve TDDP} \label{alg:PSO-IACS-TDDP}
    \SetKwInOut{Input}{Input}%
    \Input{Input parameters in Alg. \ref{alg:acs-sop};\:Number of particles $N_{\text{ptcl}}$;\: Number of iterations $N_{\text{ptcl}}^{\text{PSO}}$;\\
    Maximum number of no improvement $N_{\text{impr}}^{\text{PSO}}$;\: Improvement tolerance $\varepsilon^{\text{PSO}}$;\\
    Acceleration constants $C_1, C_2$;\: Minimum and maximum inertia weight $\omega_{\min}, \omega_{\max}$.}
    \For{$i=1$ to $N_{\mathrm{ptcl}}$}{
        Initialize the position list with random waypoints generated based on each SZ\;
        Initialize a zero velocity list with the same length as the position list\;
        Apply IACS to solve OP with the position list, record the returned prize $\mathscr{P}$, cost $\mathscr{C}_t$, and path sequence $\mathbf{S}$\;
        Update individual best-so-far prize $\mathscr{P}^{ib}$, cost $\mathscr{C}_t^{ib}$, path sequence $\mathbf{S}^{ib}$, and position list\;
    }
    Update the local-best particle, i.e., the particle with maximum $\mathscr{P}^{ib}$ or minimum $\mathscr{C}_t^{ib}$ (if $\mathscr{P}_{\max}^{ib}$ is same)\;
    Use the local-best particle to update the global best-so-far particle's $\mathscr{P}^{gb}$, $\mathscr{C}_t^{gb}$, $\mathbf{S}^{gb}$, and position list\;
    Set no improvement counter to $0$\;
    \For{$n_{it} = 1$ to $N_{\mathrm{iter}}^{\mathrm{PSO}}$}{
        \If{no improvement counter $\geq N_{\mathrm{impr}}^{\mathrm{PSO}}$}{ Break the loop\; }
        Update the inertia $\omega(n_{it})$ by Eq. \eqref{eq:PSO-LDIW}\;
        \For{each particle $k$}{
            Generate random number $R_1, R_2$\;
            \For{$j = 1$ to $M$}{
                Update the velocity of next iteration $\mathcal{V}^k_j (n_{it} + 1)$ by Eq. \eqref{eq:PSO-velocity-update}\;
                $\mathcal{V}^k_j (n_{it} + 1) \leftarrow \min \big( \mathcal{V}^k_j(n_{it} + 1),\: \mathcal{V}_j^{\max} \big)$ by Eq. \eqref{eq:PSO-vmax}\; 
                Update the position of next iteration $\big(x^k_j (n_{it} + 1),\: y^k_j (n_{it} + 1) \big)$ by Eq. \eqref{eq:PSO-position-update}\;
                Restrict $\big(x^k_j (n_{it} + 1),\: y^k_j (n_{it} + 1) \big)$ to the boundary of the SZ $\Omega_j$ if it is infeasible\;
            }
            Apply IACS to solve OP with the position list of this particle, record the returned $\mathscr{P}$, $\mathscr{C}_t$, and $\mathbf{S}$\;
            \If{$\big( \mathscr{P}^k > \mathscr{P}^{ib} \big)$ \textbf{or} $\big(\mathscr{P}^k = \mathscr{P}^{ib}$ \textbf{and } $\mathscr{C}^k_t < \mathscr{C}_t^{ib} \big)$}{
                Update individual best-so-far $\mathscr{P}^{ib}$, $\mathscr{C}_t^{ib}$, $\mathbf{S}^{ib}$, and position list\;
            }
        }
        Update the local-best particle\;
        \If{\big( $\mathscr{P}^{lb} > \mathscr{P}^{gb} + \varepsilon^{\text{PSO}}$ \big) \textbf{or} \big($\mathscr{P}^{lb} = \mathscr{P}^{gb}$ \textbf{and } $\mathscr{C}_t^{lb} < \mathscr{C}_t^{gb} - \varepsilon^{\text{PSO}}$ \big)}{
            Update the global best-so-far particle's $\mathscr{P}^{gb}$, $\mathscr{C}_t^{gb}$, $\mathbf{S}^{gb}$, and position list\;
        }
        \Else{ No improvement counter +1\; }
    }
    \Return The global best-so-far $\mathscr{P}^{gb}$, $\mathscr{C}_{t}^{gb}$, $\mathbf{S}^{gb}$, and position list.
\end{algorithm}

\subsubsection{Particle Swarm Optimization for continuous position optimization}
PSO, originally introduced in \citep{kennedy1995particle}, is a metaheuristic inspired by bird flocking behavior. In the context of CEOP-$\mathcal{N}$, PSO represents a solution as a sequence of moving waypoints confined within the valid boundaries of neighborhoods. We use $n_{it}$ to denote the iteration number. For the particle $k$, the position of its waypoint at path index $i$ during iteration $n_{it}$ is denoted as $\big(\: x^k_i(n_{it}),\: y^k_i(n_{it})\:\big)$. In PSO, the velocity of the next iteration $\big(\:\mathcal{VX}_i^k(n_{it}+1),\: \mathcal{VY}_i^k(n_{it}+1)\:\big)$ determines the motion of waypoints. The velocity of a waypoint at path index $i$ is affected by the inertia weight $\omega$, individual best-so-far (ib) position $\big( x^{ib}_i,\: y^{ib}_i \big)$, and global best-so-far (gb) position $\big( x^{gb}_i,\: y^{gb}_i \big)$, which can be updated as:
\begin{equation} \label{eq:PSO-velocity-update}
\begin{split}
    \big(\: \mathcal{VX}^k_i & (n_{it}+1),\: \mathcal{VY}^k_i(n_{it}+1) \:\big)\: = \: \omega(n_{it} + 1)\: \big(\: \mathcal{VX}^k_i(n_{it}),\: \mathcal{VY}^k_i(n_{it}) \:\big) \\&+\: C_1 R_1 \big(\: x_i^{ib}(n_{it}) - x^k_i(n_{it}),\: y_i^{ib}(n_{it}) - y^k_i(n_{it})\:\big) \:+\: C_2 R_2 \big(\: x_i^{gb}(n_{it}) - x^k_i(n_{it}),\: y_i^{gb}(n_{it}) - y^k_i(n_{it}) \:\big)
\end{split}
\end{equation}
where $C_1$ and $C_2$ are the acceleration constants for the individual best-so-far and global best-so-far solutions; $R_1$ and $R_2$ are two random variables generated from a uniform distribution ranging in $[0, 1]$. We adopt the Linear Decreasing Inertia Weight (LDIW) strategy, which leads to the lowest errors among all inertia weight strategies presented in \citep{marini2015particle}, to determine $\omega$ at each iteration:
\begin{equation} \label{eq:PSO-LDIW}
    \omega(n_{it}+1) = \omega_{\max} - \frac{\omega_{\max} - \omega_{\min}}{N_{\text{iter}}^{\text{PSO}}} \cdot n_{it} 
\end{equation}
In addition to LDIW, additional velocity restrictions are applied according to the geometric property of the SZ. Given the SZ $\Omega_k$ and its vertices $\{ v'_i, ... \}$, we define the magnitude of maximum velocity for this SZ as:
\begin{equation} \label{eq:PSO-vmax}
    \mathcal{VX}_i^{\max} = \mathcal{VY}_i^{\max} = \begin{dcases}
        r_j \quad(\text{for } c_j \in \Omega_k)\;,  & \text{if } |\Omega_k| = 1\\
        \min \Big\{ \mathscr{C}_d\big( v'_{\text{ctr}} (\Omega_k), \: \overline{v'_i v'_j} \big) \Big\} \quad(\text{for } i, j = 1, ..., |\Omega_k|, i \neq j)\; ,  & \text{otherwise}
    \end{dcases} 
\end{equation}
where $\mathscr{C}_d\big(v'_{\text{ctr}} (\Omega_k), \: \overline{v'_i v'_j}\big)$ is the shortest distance from the SZ center to the line segment $\overline{v'_i v'_j}$, where $\overline{v'_i v'_j}$ is the edge of the polygon formed by SZ vertices. Note that we set the initial waypoint position as a random point located in a circle with center $v'_{\text{ctr}} (\Omega_k)$ and radius $r = \min \Big\{ \mathscr{C}_d\big( v'_{\text{ctr}} (\Omega_k), \: \overline{v'_i v'_j} \big) \Big\}$. As suggested by \citet{engelbrecht2012particle}, the velocity vector should be initialized as a zero vector. The position at path index $i$ can be then updated as:
\begin{equation} \label{eq:PSO-position-update}
    \big(\: x_i^k(n_{it}+1),\: y_i^k(n_{it}+1)\:\big) = \big(\: x_i^k(n_{it}) + \mathcal{VX}_i^k(n_{it}+1),\: y_i^k(n_{it}) + \mathcal{VY}_i^k(n_{it}+1)\:\big)
\end{equation}
To guarantee the feasibility of the solution, each waypoint is confined to the boundary of the corresponding SZ if its next position falls outside of the SZ. The algorithm establishes a line segment connecting the current and next positions. It subsequently evaluates each circle covered by the SZ to identify the precise boundary point, i.e., the intersection point where the line segment intersects a specific circle. This boundary point can be uniquely determined due to the convex nature of the SZ.

\subsubsection{Inherited Ant Colony System for path sequence optimization}
Due to maximum velocity constraints, alterations to waypoint positions between consecutive iterations may be restricted. Consequently, the global-best solution obtained in the previous iteration often remains high quality, and this prior knowledge can be leveraged to enhance the initialization phase. In IACS, the nearest neighbor heuristic is only used to initialize the pheromone matrix in the first iteration. IACS then `inherits' the global-best ant from the previous iteration to calculate the initial pheromone:
\begin{eqnarray} \label{eq:IACS-tau0}
    \tau_0(n_{it}) = \dfrac{\mathscr{P}^{gb} (n_{it} - 1)}{\mathscr{C}_{f}^{gb} (n_{it} - 1)},\quad\quad n_{it} > 1
\end{eqnarray}
An additional global updating rule is applied to reinforce the impact of the previous global-best solution. Note that in CEOP-$\mathcal{N}$, IACS is used as the solver for OP instead of SOP because PSO can provide a unique waypoint of each SZ to IACS. 

\section{Experiments, Results and Discussion} \label{sec:experiments-and-results}
This section investigates the advantages of incorporating overlapped neighborhoods, as revealed through a series of computational experiments conducted on instances of CEOP and TDDP. Our algorithm was implemented using the C++20 standard, and experiments were executed on a Linux machine with AMD EPYC 7742 (2.25 GHz and 8 GB RAM). To assist with understanding the CRaSZe-AntS architecture and flow, we provide a graphical illustration and supplementary materials, available online via the GitHub link in \autoref{fn-github}. Notably, all CEOP and TDDP instances used were derived from CETSP instances presented in \citet{mennell2009heuristics}. \citet{yang2018double} examined the performance of their algorithm through 10 runs, however, in this work, we increase this to 20 individual executions with a view to examining additional repeatability\footnote[2]{All experiment data, including solution paths with waypoint positions, discretization layouts, and result summary of CRaSZe-AntS and benchmark algorithms on SOP, CEOP, and TDDP instances are provided as supplementary materials; available online via Github link in \autoref{fn-github}.}. The results presented in the subsequent tables represent the mean and Standard Deviations of outcomes obtained from 20 independent executions, where we denote them as `avg.' and `SD', respectively. The following attributes were considered when evaluating the solution performance:
\begin{itemize}
    \item[$\mathscr{t}_\text{alg}$] Algorithm execution time (seconds).
    \item[$\mathscr{P}$] Total collected prize of the path.
    \item[$\mathscr{C}_d$] Total cost of the path in distance metric.
    \item[$\mathscr{C}_t$] Total cost of the path in time metric.
\end{itemize}

\begin{table}[!t]
    \centering
    \begin{threeparttable}[b]
        \caption{Scores achieved with varying ACS settings} \label{tab:score-acs-setting}
        \begin{tabular}{lllll}
            \toprule
            \diagbox{$N_{\text{ant}}$}{$N_{\text{iter}}$} & 125 & 250 & 500 & 1000\\
            \hline
            10 & 4.573 & 5.781 & 6.085 & 5.766\\
            20 & 6.161 & 6.214 & 6.305 & 5.743\\
            40 & 6.733 & \textbf{7.031} & 6.143 & 5.050\\
            80 & 6.799 & 6.932 & 6.081 & 4.198\\
            \bottomrule
        \end{tabular}
        \scriptsize
        \parbox[b]{9cm}{The \textbf{bold} value indicates the best result.}
    \end{threeparttable}
\end{table}

\subsection{CEOP with uniform neighborhoods}
The significance of incorporating overlapped neighborhoods becomes evident when highlighting the prize attribute in the context of CEOP, enabling the simultaneous collection of adjacent prizes. To illustrate this concept further, we explore how RSZD discretizes CEOP into SOP, how ACS (in \S \ref{sec:ceop-acs-arc}) solves SOP, and how the arc search algorithm optimizes the waypoints' positions.

\begin{table*}[!b]
\setlength\tabcolsep{0pt}
\setlength\extrarowheight{2pt}
\begin{threeparttable}[b]
\caption{Performance summary of IPD-ACS and RSZD-ACS (first discretize the CEOP and then solve the SOP).} \label{tab:sop}
\begin{tabular*}{\columnwidth}{@{\extracolsep{\fill}}*{14}{l}}
\toprule
\multirow{3}{*}{Instances} & \multirow{3}{*}{$\mathscr{B}_d$} & \multicolumn{6}{l}{IPD-ACS} & \multicolumn{6}{l}{RSZD-ACS} \\ \cline{3-8} \cline{9-14} 
 &  & \multicolumn{2}{l}{$\mathscr{t}_\text{alg}$ (seconds)} & \multicolumn{2}{l}{$\mathscr{C}_d$} & \multicolumn{2}{l}{$\mathscr{P}$} & \multicolumn{2}{l}{$\mathscr{t}_\text{alg}$ (seconds)} & \multicolumn{2}{l}{$\mathscr{C}_d$} & \multicolumn{2}{l}{$\mathscr{P}$} \\ \cline{3-4}\cline{5-6}\cline{7-8} \cline{9-10}\cline{11-12}\cline{13-14}
  &  & avg. & SD & avg. & SD & avg. & SD & avg. & SD & avg. & SD & avg. & SD \\ \midrule
\textit{bubbles1} & 314.22 & 1.72 & 0.03 & 313.84 & 0 & 312 & 0 & \textbf{0.05} & 0 & 298.26 & 0 & \textbf{348} & 0 \\
\textit{bubbles2} & 385.45 & 10.24 & 3.14 & 383.98 & 1.17 & 514.4 & 7.09 & \textbf{0.21} & 0.07 & 381.59 & 2.36 & \textbf{640.6} & 5.26 \\
\textit{bubbles3} & 476.96 & 26.51 & 7.61 & 475.86 & 0.92 & 639.3 & 18.8 & \textbf{1.1} & 0.42 & 474.44 & 1.77 & \textbf{842.6} & 11.58 \\
\textit{bubbles4} & 724.91 & 66.23 & 20.38 & 722.48 & 2.13 & 886.7 & 7.68 & \textbf{2.76} & 1.12 & 721.79 & 1.97 & \textbf{1121.8} & 15.05 \\
\textit{bubbles5} & 934.34 & 114.75 & 33.8 & 932.68 & 1.08 & 1024.1 & 11.58 & \textbf{4.02} & 1.14 & 930.82 & 2.73 & \textbf{1273.2} & 7.98 \\
\textit{bubbles6} & 1106.69 & 204.07 & 61.13 & 1104.51 & 1.43 & 1133.9 & 18.32 & \textbf{6.15} & 1.97 & 1103.57 & 2.02 & \textbf{1356.1} & 6.11 \\
\textit{bubbles7} & 1446.58 & 353.65 & 91.28 & 1444.15 & 1.92 & 1307.45 & 13.01 & \textbf{8.75} & 4.2 & 1441.86 & 4.29 & \textbf{1554.85} & 13.13 \\
\textit{bubbles8} & 1752.05 & 860.68 & 344.38 & 1750.77 & 0.77 & 1846.65 & 37.17 & \textbf{17.56} & 9.22 & 1747.53 & 2.35 & \textbf{2250.9} & 24.38 \\
\textit{bubbles9} & 2033.3 & 1262.54 & 510.09 & 2032.03 & 1.13 & 2822.2 & 35.3 & \textbf{23.9} & 17.81 & 2028.87 & 3.45 & \textbf{3384.55} & 20.44 \\ \midrule
\textit{bubbles1} & 209.48 & 1.2 & 0.01 & 206.47 & 0.55 & 168 & 0 & \textbf{0.04} & 0 & 192.12 & 0 & \textbf{192} & 0 \\
\textit{bubbles2} & 256.97 & 6.12 & 1.42 & 255.64 & 0.85 & 306.2 & 4.47 & \textbf{0.14} & 0.03 & 253.98 & 2.5 & \textbf{382.8} & 4.92 \\
\textit{bubbles3} & 317.98 & 16.4 & 5.55 & 316.53 & 1.11 & 392.6 & 6.3 & \textbf{0.72} & 0.23 & 314.53 & 2.44 & \textbf{520.5} & 6.57 \\
\textit{bubbles4} & 483.28 & 49.63 & 13.99 & 481.62 & 1.16 & 643.2 & 22.02 & \textbf{1.55} & 0.57 & 479.47 & 3.12 & \textbf{809} & 8.16 \\
\textit{bubbles5} & 622.9 & 88.61 & 20.25 & 621.55 & 1.18 & 807.8 & 16.83 & \textbf{3.15} & 1.22 & 618.14 & 2.46 & \textbf{995.4} & 13.93 \\
\textit{bubbles6} & 737.8 & 162.54 & 60.77 & 736.47 & 1.4 & 885.7 & 12.24 & \textbf{6.29} & 3.96 & 731.72 & 4.57 & \textbf{1128.9} & 17.35 \\
\textit{bubbles7} & 964.39 & 238.62 & 57.1 & 962.46 & 1.4 & 1036 & 17.23 & \textbf{8.7} & 4.9 & 957.11 & 4.68 & \textbf{1287.1} & 10.94 \\
\textit{bubbles8} & 1168.03 & 483.18 & 209.57 & 1166.54 & 1.1 & 1264.7 & 29.63 & \textbf{13.02} & 6.8 & 1164.08 & 3.17 & \textbf{1579.15} & 19.04 \\
\textit{bubbles9} & 1355.53 & 718.75 & 220.87 & 1355 & 0.47 & 1973.35 & 39.46 & \textbf{17.33} & 7.26 & 1352.6 & 2.35 & \textbf{2459.35} & 24.73 \\ \midrule
\textit{bubbles1} & 104.74 & 0.49 & 0 & 103.73 & 0.44 & 48 & 0 & \textbf{0.02} & 0 & 94.86 & 7.77 & \textbf{52.8} & 5.88 \\
\textit{bubbles2} & 128.48 & 2.32 & 0.38 & 125.57 & 0.25 & 84 & 0 & \textbf{0.05} & 0.01 & 123.66 & 5.34 & \textbf{106.8} & 8.77 \\
\textit{bubbles3} & 158.99 & 5.78 & 0.99 & 158.22 & 0.58 & 129 & 2.49 & \textbf{0.23} & 0.06 & 156.87 & 0.04 & \textbf{176} & 0 \\
\textit{bubbles4} & 241.64 & 17.98 & 6.18 & 240.62 & 1.01 & 268.2 & 5.86 & \textbf{0.53} & 0.17 & 238.97 & 1.93 & \textbf{339.8} & 4.47 \\
\textit{bubbles5} & 311.45 & 40.37 & 12.18 & 310.5 & 0.95 & 385.7 & 6.21 & \textbf{0.97} & 0.32 & 307.5 & 2.36 & \textbf{481.6} & 3.67 \\
\textit{bubbles6} & 368.9 & 68.19 & 20.04 & 367.75 & 1.02 & 469.2 & 11.43 & \textbf{1.8} & 0.76 & 365.76 & 2.53 & \textbf{596.2} & 6.87 \\
\textit{bubbles7} & 482.19 & 129.26 & 41.56 & 480.91 & 1.14 & 641.1 & 24.92 & \textbf{3.83} & 1.55 & 477.35 & 4.36 & \textbf{805.9} & 11.97 \\
\textit{bubbles8} & 584.02 & 190.93 & 55.01 & 582.62 & 0.99 & 760.2 & 18.93 & \textbf{5.86} & 3.35 & 577.92 & 4.11 & \textbf{941.9} & 11.99 \\
\textit{bubbles9} & 677.77 & 285.4 & 83.29 & 676.94 & 1.36 & 877.7 & 8.61 & \textbf{7.83} & 3.64 & 675.54 & 1.86 & \textbf{1104.7} & 12.41 \\
\bottomrule
\end{tabular*}
\scriptsize
\vspace*{0.1cm}
\parbox[b]{\textwidth}{For an OP-related solution, a solution's quality is higher than another when it achieves a higher prize or a lower cost when the prizes are the same. The layout from RSZD outperforms IPD's layout across all instances tested, numerically with an averaged $\mathscr{t}_{\text{alg}}$ decrease of $-96.95\,\%$ and $\mathscr{P}$ increase of $24.03\,\%$. Furthermore, employing RSZD makes the ACS yield more robust solutions because CRaSZe-AntS has significantly reduced SD for $\mathscr{t}_{\text{alg}}$ and $\mathscr{P}$ across \textit{bubbles1-9}.}
\end{threeparttable}
\end{table*}

\subsubsection{ACS parameter setting} \label{sec:acs-param}
Because ACS was initially proposed to tackle TSP \citep{dorigo1997ant}, we first examine the performance of ACS in CETSP instances using different parameters. As suggested by \citet{dorigo1997ant}, we set heuristic importance factor $\beta = 2$, pheromone evaporation rate $\alpha = \rho = 0.1$; exploitation probability $q_0 = 0.9$, and initial pheromone $\tau_0 = (\mathscr{C}_d^{nn})^{-1}$, where $\mathscr{C}_d^{nn}$ is the distance cost of the path yielded from nearest neighbor heuristic \citep{rosenkrantz1977analysis}. Sixteen pairs of settings, i.e., $N_{\text{ant}} \in \big\{ 10, 20, 40, 80 \big\}$ and $N_{\text{iter}}^{\text{ACS}} \in \big\{ 125, 250, 500, 1000 \big\}$, were tested in \textit{bubbles1-9}. For example, setting 1 refers to $N_{\text{ant}} = 10, N_{\text{iter}}^{\text{ACS}} = 125$; Setting 2 refers to $N_{\text{ant}} = 10, N_{\text{iter}}^{\text{ACS}} = 250$, etc. Note that \textit{bubbles1-9} are CETSP instances in \citep{mennell2009heuristics}, characterized by well-organized neighborhood structures with sufficiently substantial overlaps. Though other CETSP instances, such as \textit{rotatingDiamonds} and \textit{concentricCircles} series, also possess structured layouts, their complexity is lower than that of \textit{bubbles} series when constructing SZ layout. For instance, most SZs in \textit{rotatingDiamonds} and \textit{concentricCircles} are consecutive SZs of degree 2. Moreover, \textit{bubbles} series exhibits a progressively increasing computational complexity from 1 to 9, which helps to discern the performance difference among evaluated algorithms. To select a general setting for later experiments, we define the score of setting $i$ based on path distance cost and computational time:
\begin{eqnarray} \label{eq:acs-score}
score(i) = \sum_{\textit{bubbles}} 0.5 \cdot \bigg( \dfrac{\max_{i} \big\{ \mathscr{C}_d (i) \big\} - \mathscr{C}_d(i)}{\max_{i} \big\{ \mathscr{C}_d (i) \big\} - \min_{i} \big\{\mathscr{C}_d (i) \big\} } + \dfrac{\max_{i} \big\{ \mathscr{t}_{\text{alg}}(i) \big\} - \mathscr{t}_{\text{alg}}(i)}{ \max_i \big\{ \mathscr{t}_{\text{alg}}(i) \big\} - \min_i \big\{ \mathscr{t}_{\text{alg}}(i) \big\} }  \bigg), \quad i = 1, 2, ..., 16
\end{eqnarray}
where $i$ refers to the case number of \textit{bubbles}; $\max_{i} \big\{ \mathscr{C}_d (i) \big\}, \min_{i} \big\{ \mathscr{C}_d (i) \big\}$ and $\max_i \big\{ \mathscr{t}_{\text{alg}}(i) \big\}, \min_i \big\{ \mathscr{t}_{\text{alg}}(i) \big\}$ are the maximum (and minimum) achieved path distance cost and computational time among all settings within each \textit{bubbles} instance. Table ~\ref{tab:score-acs-setting} summarized the scores of all settings, which indicates setting 10 (i.e., $N_{\text{ant}} = 40, N_{\text{iter}}^{\text{ACS}} = 250$) obtained the highest scores among all instances. Therefore, in all experiments of the following sections, except when indicated differently, $N_{\text{ant}}$ and $N_{\text{iter}}^{\text{ACS}}$ were set to 40 and 250, respectively. Additionally, we established a minimum improvement tolerance as $\varepsilon^{\text{ACS}} = 10^{-4}$, which means ACS would terminate if the fitness difference is less than $\varepsilon^{\text{ACS}}$ for several iterations. To balance the computation time and solution quality, we allow a maximum number of no improvements as $N_{\text{iter}}^{\text{ACS}} / \:10 = 25$. 

\subsubsection{The performance of discretization schemes}
To illustrate the computational and strategic distinctions between employing single-neighborhood versus overlapped-neighborhood strategies for discretizing the CEOP, our RSZD was compared with the IPD scheme proposed by \citet{carrabs2017novel}. The IPD scheme is selected for comparison because it represents one of the few strategies that effectively discretizes CETSP into the GTSP, and this can be mapped well to the CEOP scenario. Moreover, as evidenced in \citep{carrabs2017novel}, IPD outperforms the perimetral discretization scheme proposed in \citep{behdani2014integer}. We set the iteration number $N_{\text{iter}}^{\text{RSZD}} = 10$. The selection of this value is determined by tests involving a range of $[5, 10, 15, 20]$. The performance difference is not significant, mainly because we are testing on organized instances (i.e.,~\textit{bubbles} series). The budget level for each instance is set to $90\,\%$, $60\,\%$, and $30\,\%$ of the best-known cost of its respective CETSP instance~\citep{di2022genetic}. For example, the budget $\mathscr{B}_d$ for \textit{bubbles1} is set to $314.22$, $209.48$ and $104.74$, respectively (the best-known cost is $349.13$). Moreover, we used three discrete points to represent a single circle, keeping the same configuration as \citep{carrabs2020adaptive}.

\begin{figure*}[!t]
    \centering
    \begin{subfigure}{0.9\textwidth} 
        \centering
        \includegraphics[width=\textwidth]{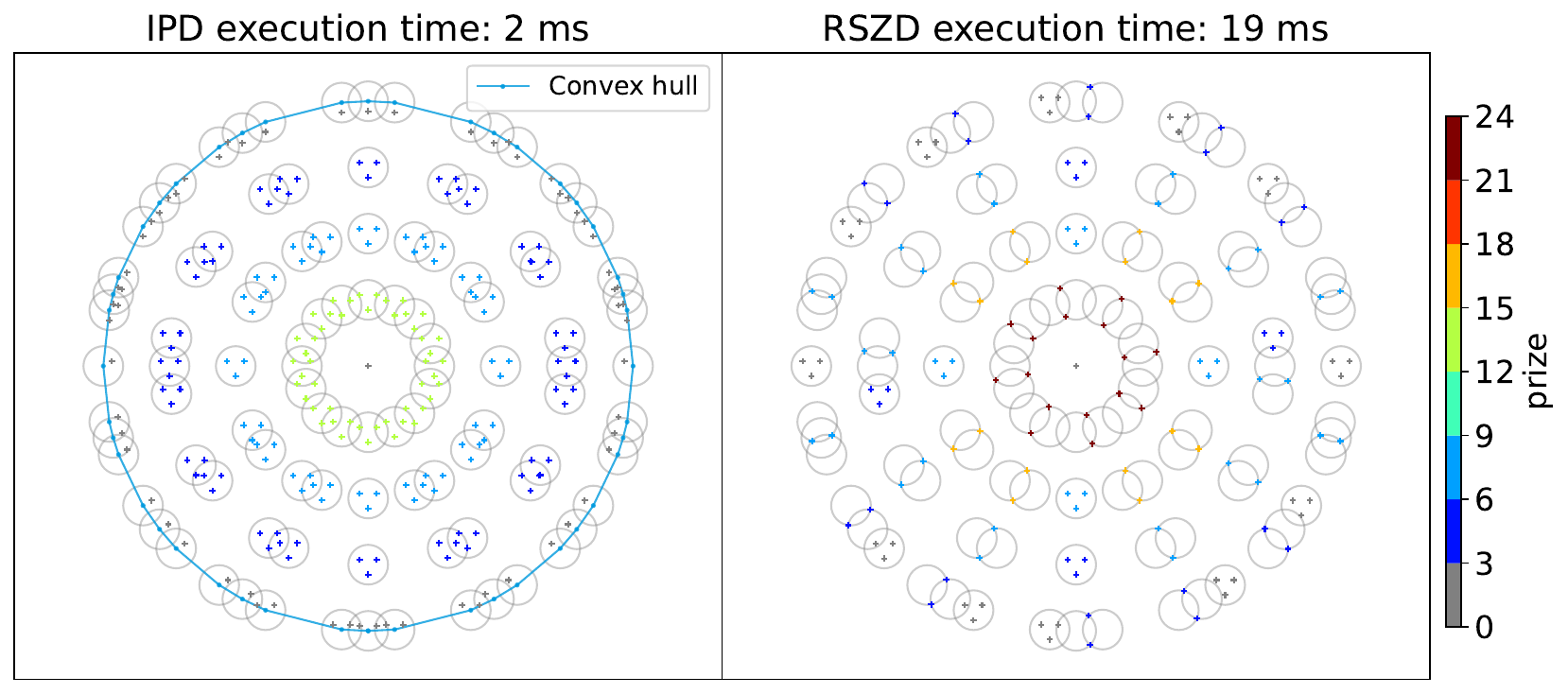}
        \caption{\textit{concentricCircles4}.}
        \label{fig:concentricCircles4-disc}
    \end{subfigure}
    \begin{subfigure}{0.9\textwidth} 
        \centering
        \includegraphics[width=\textwidth]{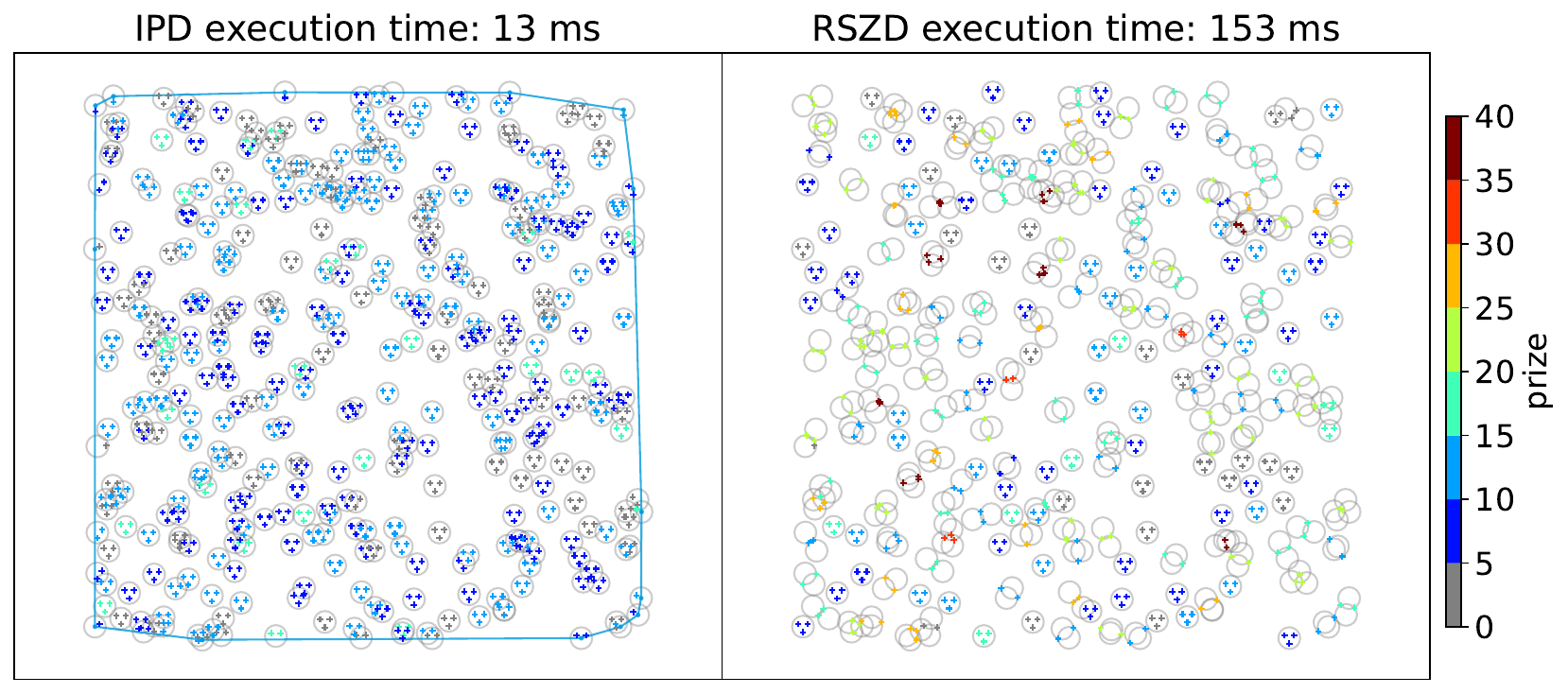}
        \caption{\textit{rd400} with overlap ratio $\phi_{or} = 0.02$.}
        \label{fig:rd400-dis}
    \end{subfigure}
    \caption{Discretization results of IPD and RSZD. RSZD and IPD have the same worst-time complexity as $\mathcal{O}(N)$, where $N$ is the number of target circles. Here RSZD needs $\sim$10 times slower computation time than IPD because we set $N_{\text{iter}}^{\text{RSZD}} = 10$.} \label{fig:IPD-vs-RSZD}
\end{figure*}

Table \ref{tab:sop} shows the results achieved by ACS when utilizing the waypoint layouts generated by RSZD and IPD. RSZD's layout exhibited superior computational efficiency and solution quality performance across all tested scenarios, outperforming IPD. For instance, in \textit{bubbles5} (with 250 target circles), the RSZD-ACS yielded solutions approximately 30 times faster than IPD-ACS, also with a prize increase of $24.32\,\%$. Furthermore, ACS can yield more robust solutions by employing RSZD. For example, RSZD-ACS exhibits a Standard Deviation (SD) of 17.81 seconds in computation time and 20.44 in prize collection across 20 individual executions when applied to \textit{bubbles9} with a $90\,\%$ budget level. While the SD values for IPD-ACS are substantially higher, with 510.09 seconds for computation time and 35.3 for prize collection. Note that the magnitude of `35.3' remains acceptable because targets' prizes are within a predefined range $[1, 12]$ in \textit{bubbles9}. It is important to highlight that IPD generates three times as many discrete points as ACS's input while RSZD extracts SZs to reduce the input size (at worst the same as the original graph). Furthermore, though IPD is proven to be effective for CETSP \citep{carrabs2017novel}, it exhibits less effective performance in CEOP due to its accumulated discretization error. This limitation can hinder ACS's ability to incorporate additional targets while adhering to budget constraints. Fig. \ref{fig:IPD-vs-RSZD} visualizes the discretization layouts of IPD and RSZD for \textit{concentricCircles4} and \textit{rd400} with overlap ratio $\phi_{or} = 0.02$. Note that \citet{mennell2009heuristics} proposed \textit{overlap ratio} to describe the complexity of a CETSP instance, $\phi_{or} =  r \:/\: l_{contain}$, where $l_{contain}$ is the length of the smallest square to contain all circles. We clarify the definition of overlap ratio as $\phi_{or} = r \: /\: \max \big(x_{\max} - x_{\min}, y_{\max} - y_{\min} \big)$, where $r$ is the uniform radius for all target nodes; $x_{\min}$, $x_{\max}$, $y_{\max}$, and $y_{\min}$ are the minimum and maximum 2D coordinates among all nodes (including start and end depot).

\begin{table*}[!b]
\setlength\tabcolsep{0pt}
\setlength\extrarowheight{2pt}
\begin{threeparttable}[b]
\caption{Performance summary of GRASP and CRaSZe-AntS for CEOP.} \label{tab:ceop}
\begin{tabular*}{\columnwidth}{@{\extracolsep{\fill}}*{14}{l}}
\toprule
\multirow{3}{*}{Instances} & \multirow{3}{*}{$\mathscr{B}_d$} & \multicolumn{6}{l}{GRASP} & \multicolumn{6}{l}{CRaSZe-AntS (RSZD-ACS-ARC)} \\ \cline{3-8} \cline{9-14} 
 &  & \multicolumn{2}{l}{$\mathscr{t}_\text{alg}$ (seconds)} & \multicolumn{2}{l}{$\mathscr{C}_d$} & \multicolumn{2}{l}{$\mathscr{P}$} & \multicolumn{2}{l}{$\mathscr{t}_\text{alg}$ (seconds)} & \multicolumn{2}{l}{$\mathscr{C}_d$} & \multicolumn{2}{l}{$\mathscr{P}$} \\ \cline{3-4}\cline{5-6}\cline{7-8} \cline{9-10}\cline{11-12}\cline{13-14}
  &  & AVG & SD & AVG & SD & AVG & SD & AVG & SD & AVG & SD & AVG & SD \\ \midrule
\textit{bubbles1} & 314.22 & \textbf{0.02} & 0 & 308.25 & 0.94 & \textbf{357.6} & 4.8 & 0.2 & 0 & 295.82 & 0.07 & 348 & 0 \\
\textit{bubbles2} & 385.45 & \textbf{0.36} & 0.16 & 382.57 & 1.51 & \textbf{652} & 16 & 0.75 & 0.26 & 378.64 & 2.91 & 640.2 & 5.29 \\
\textit{bubbles3} & 476.96 & \textbf{2.43} & 0.65 & 475.63 & 1.01 & 879.6 & 28.07 & 4.52 & 1.68 & 471.39 & 2.79 & \textbf{879.7} & 8.08 \\
\textit{bubbles4} & 724.91 & \textbf{8.57} & 2.79 & 722.86 & 1.13 & 1154 & 17.62 & 10.67 & 4.05 & 720.56 & 2.81 & \textbf{1157.4} & 8.99 \\
\textit{bubbles5} & 934.34 & 20.95 & 5.29 & 931.82 & 1.29 & 1298.3 & 10.92 & \textbf{17.05} & 7.31 & 929.36 & 2.47 & \textbf{1305.6} & 5.99 \\
\textit{bubbles6} & 1106.69 & 63.44 & 13.89 & 1104.64 & 0.87 & \textbf{1379.5} & 9.13 & \textbf{24.71} & 10.84 & 1101.94 & 2.14 & 1378.3 & 3.54 \\
\textit{bubbles7} & 1446.58 & 130.52 & 29.68 & 1445.34 & 0.73 & 1589.35 & 16.09 & \textbf{50.72} & 26.67 & 1439.14 & 3.14 & \textbf{1607.55} & 12.36 \\
\textit{bubbles8} & 1752.05 & 332.82 & 88.75 & 1751.33 & 0.46 & 2262.6 & 37.34 & \textbf{84.2} & 34.64 & 1745.95 & 3.19 & \textbf{2322} & 5.37 \\
\textit{bubbles9} & 2033.3 & 790.79 & 180.68 & 2032.37 & 0.71 & 3455.5 & 30.35 & \textbf{175.82} & 75.68 & 2026.14 & 4.62 & \textbf{3463.9} & 10.46 \\ \midrule
\textit{bubbles1} & 209.48 & \textbf{0.01} & 0.01 & 203.84 & 2.49 & 205.8 & 4.28 & 0.14 & 0 & 207.89 & 0.05 & \textbf{216} & 0 \\
\textit{bubbles2} & 256.97 & \textbf{0.22} & 0.08 & 254.33 & 2 & 372.8 & 23.28 & 0.53 & 0.17 & 251.55 & 2.54 & \textbf{380.8} & 3.49 \\
\textit{bubbles3} & 317.98 & \textbf{1.01} & 0.45 & 316.73 & 1.18 & 521.2 & 20.27 & 2.11 & 0.7 & 312.65 & 2.73 & \textbf{533.4} & 4.78 \\
\textit{bubbles4} & 483.28 & \textbf{5.39} & 1.89 & 481.7 & 1.09 & \textbf{886.3} & 24.39 & 7.22 & 3.05 & 478.08 & 3.08 & 825.6 & 4.88 \\
\textit{bubbles5} & 622.9 & 15.14 & 4.6 & 621.18 & 1.12 & \textbf{1063.2} & 19.94 & \textbf{11.59} & 4.26 & 617.84 & 2.57 & 1027.5 & 11.36 \\
\textit{bubbles6} & 737.8 & 32.75 & 10.18 & 736.21 & 1.15 & 1156.3 & 26.36 & \textbf{22.45} & 8.86 & 730.16 & 3.95 & \textbf{1161.7} & 7.3 \\
\textit{bubbles7} & 964.39 & 87.1 & 18.98 & 962.98 & 0.89 & 1315.05 & 16.12 & \textbf{38.66} & 16.13 & 957.79 & 4.66 & \textbf{1319.5} & 7.21 \\
\textit{bubbles8} & 1168.03 & 272.39 & 61.14 & 1167.42 & 0.37 & 1631.7 & 46.48 & \textbf{48.2} & 22.25 & 1162.4 & 3.31 & \textbf{1639.4} & 14.54 \\
\textit{bubbles9} & 1355.53 & 479.8 & 131.5 & 1355.09 & 0.31 & \textbf{2576.6} & 99.28 & \textbf{62.58} & 24.45 & 1349.69 & 2.35 & 2558.35 & 14.6 \\ \midrule
\textit{bubbles1} & 104.74 & \textbf{0} & 0 & 98.21 & 2.06 & 51.6 & 9.37 & 0.07 & 0.02 & 103.54 & 3.66 & \textbf{59.4} & 2.62 \\
\textit{bubbles2} & 128.48 & \textbf{0.01} & 0.01 & 125.66 & 2.12 & \textbf{109.2} & 6.08 & 0.18 & 0.01 & 121.08 & 5.66 & 103.4 & 7.07 \\
\textit{bubbles3} & 158.99 & \textbf{0.1} & 0.08 & 156.4 & 2.26 & \textbf{181.9} & 7.22 & 0.87 & 0.23 & 156.62 & 0.02 & 176 & 0 \\
\textit{bubbles4} & 241.64 & \textbf{0.66} & 0.36 & 239.62 & 1.72 & \textbf{349.6} & 24.06 & 2.24 & 0.62 & 237.91 & 1.7 & 345 & 6.56 \\
\textit{bubbles5} & 311.45 & \textbf{2.16} & 0.85 & 309.97 & 1.16 & \textbf{499} & 21.07 & 3.52 & 0.97 & 306.27 & 1.03 & 484.8 & 2.4 \\
\textit{bubbles6} & 368.9 & \textbf{4.91} & 2.2 & 367.81 & 0.73 & \textbf{629.5} & 25.4 & 6.4 & 2.02 & 362 & 2.35 & 601 & 5.2 \\
\textit{bubbles7} & 482.19 & 18.03 & 7.9 & 481.16 & 0.78 & \textbf{869.3} & 27.25 & \textbf{14.24} & 5.47 & 475.18 & 2.63 & 821.4 & 7.3 \\
\textit{bubbles8} & 584.02 & 47.15 & 15.25 & 583.32 & 0.63 & 941.65 & 32.93 & \textbf{27.59} & 8.82 & 578.38 & 4.19 & \textbf{983.3} & 9.21 \\
\textit{bubbles9} & 677.77 & 89.95 & 29.43 & 677.26 & 0.44 & \textbf{1163.35} & 55.76 & \textbf{29.94} & 11.43 & 672.15 & 3.48 & 1144.55 & 15.04 \\
\bottomrule
\end{tabular*}
\scriptsize
\vspace*{0.1cm}
\parbox[b]{\textwidth}{In small-scale instances, GRASP outperforms CRaSZe-AntS because optimizing a promising candidate is efficient. While for larger-scale instances, e.g. \textit{bubbles5-9}, CRaSZe-AntS demonstrated an averaged $\mathscr{t}_{\text{alg}}$ reduction of $-40.6\,\%$ and $\mathscr{P}$ decrease of $-0.56\,\%$. Furthermore, CRaSZe-AntS also demonstrates a reduced SD in terms of $\mathscr{t}_{\text{alg}}$ and $\mathscr{P}$, compared to GRASP.}
\end{threeparttable}
\end{table*}

\subsubsection{Impact of overlapped neighborhoods on CEOP solutions} \label{sec:result-ceop}
The solution to our SOP can be regarded as an approximate solution to CEOP because the SOP solution must be a feasible CEOP solution with all waypoints being SZs' vertices. To comprehensively assess the influence of incorporating overlapped neighborhoods on CEOP solutions, we re-solve \textit{bubbles1-9}, employing a local search operator, i.e., arc search algorithm. Although RSZD draws inspiration from the SZ strategy outlined in \citep{mennell2009heuristics, wang2019steiner}, none of these algorithms have been selected as benchmark references. This decision is based on the fact that these algorithms primarily focus on generating high-quality SZ layouts, which is not the primary objective of this paper. Moreover, re-producing SZVNS \citep{wang2019steiner} is not possible due to the absence of specific technical details, particularly concerning determining the optimal waypoint position of a SZ. Also, we aim to present the advantage of considering overlapped neighborhoods as opposed to single-neighborhood strategies in CEOP and CEOP-$\mathcal{N}$ scenarios. Therefore, our comparative analysis used the Greedy Randomized Adaptive Search Procedure (GRASP) \citep{vstefanikova2020greedy} as the benchmark algorithm. The overall framework (as stated in \S \ref{sec:related-work}) of GRASP is similar to that of CRaSZe-AntS. Both strategies start by discretizing the continuous CEOP, then optimizing the sequence of waypoints and fine-tuning the continuous positions of waypoints. We summarize the principal distinctions between GRASP and CRaSZe-AntS as follows: (1) The solution construction phase of GRASP predominantly relies on local information such as distance between waypoints, the relationship between circle and line segment, etc. In contrast, ACS utilizes the pheromone matrix to represent the balance between global and local information; (2) `local search' in GRASP optimizes waypoints' positions within individual circles, while our arc search algorithm optimizes waypoints' positions within SZs. 
\begin{figure*}[!t] 
    \centering
    \begin{subfigure}{0.9\textwidth} 
        \centering
        \includegraphics[width=\textwidth]{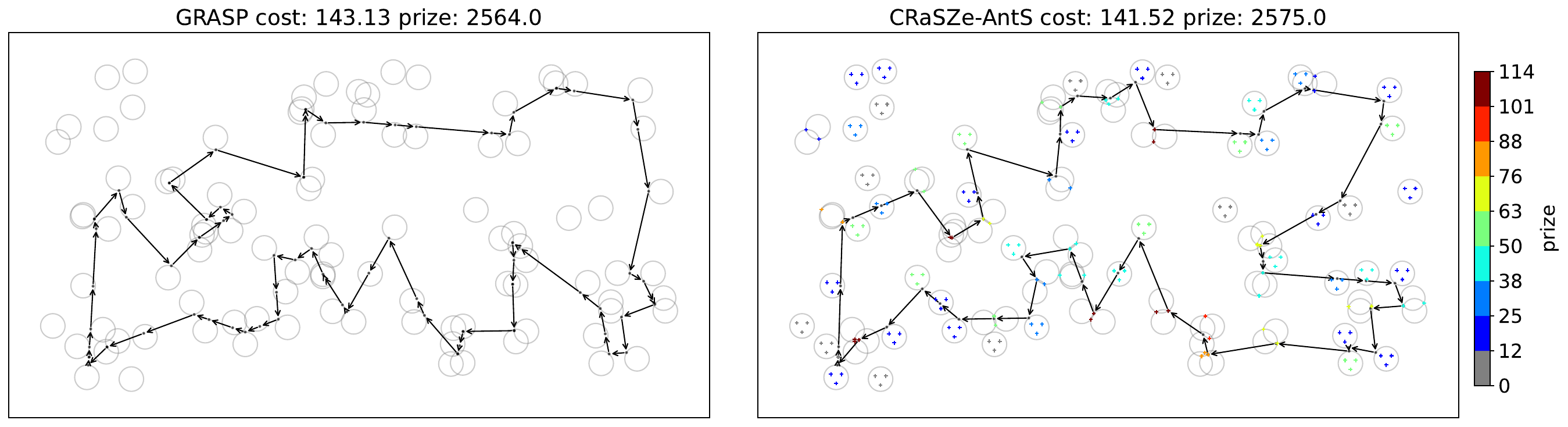}
        \caption{\textit{kroD100} with overlap ratio 0.02 and budget 143.136.} \label{fig:kroD100or002-ceop}   
    \end{subfigure}
    \begin{subfigure}{0.9\textwidth} 
        \centering
        \includegraphics[width=\textwidth]{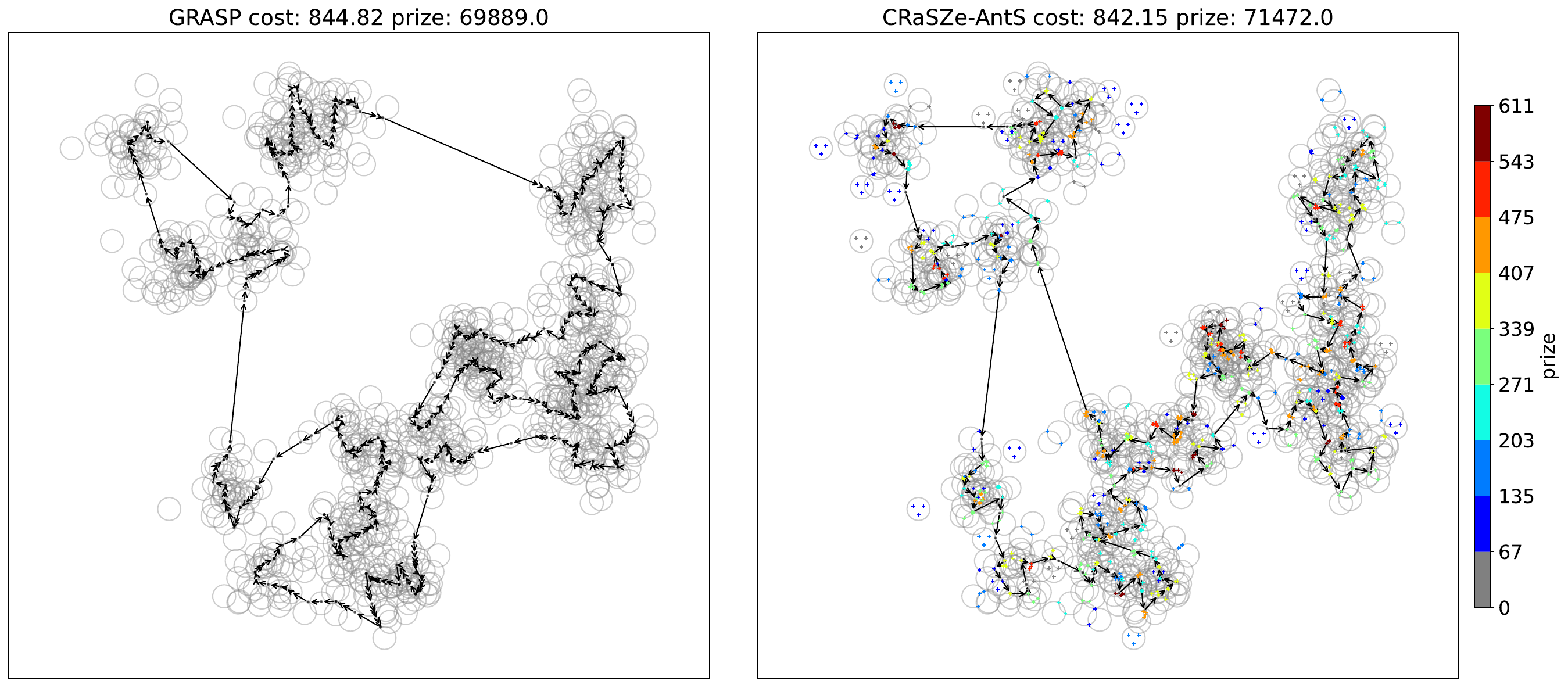}
        \caption{\textit{dsj1000} with overlap ratio 0.02 and budget 844.84.} \label{fig:dsj1000or002-ceop}
    \end{subfigure}
    \caption{Paths of GRASP and CRaSZe-AntS in solving two CEOP instances.}
    \label{fig:overlap-in-ceop}
\end{figure*}

GRASP is characterized by the operators of `add location' and `local search'. The `add location' operator extends the path by greedily selecting nodes with the lowest visitation cost until the budget constraint is met. In cases where the constructed path exceeds feasibility, GRASP iteratively removes a segment of nodes until the path becomes feasible. Whenever a feasible path is achieved, it is added to a candidate list, from which a high-quality path is randomly selected as the starting path for the subsequent iteration. The solution construction phase continues until no better path can be found. GRASP employs a local search operator to refine the global-best path after the construction phase. Initially, 2-opt is applied to improve the path sequence. Subsequently, GRASP optimizes the waypoint position for a single circle (similarly to this element of the arc search algorithm, introduced in \S \ref{sec:local-search}). Lastly, the `add location' operator is re-invoked to replace nodes in the path with potential higher-profit nodes. 

\begin{figure}[!t] 
    \centering
    \includegraphics[width=0.9\textwidth]{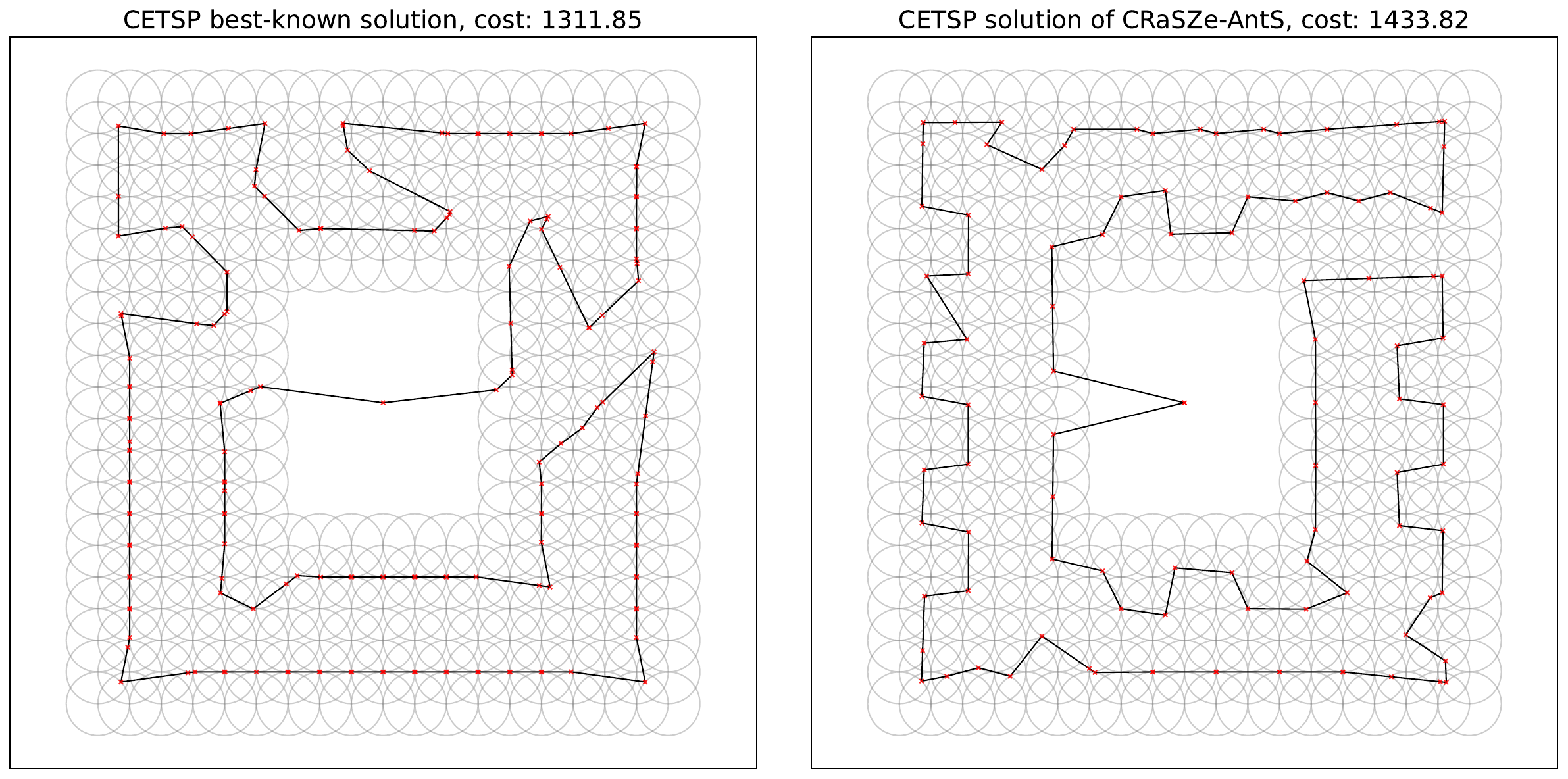}
    \caption{Paths of the best-known solution and CRaSZe-AntS in solving CETSP instance \textit{bubbles6}. Compared to CRaSZe-AntS, the best-known path exhibits a notably `straight' and `smooth' trajectory. Many SZs of degree four form the `saw-tooth' pattern in the path. Another relates to extreme scenarios, such as four circles intersecting at a single point.}
    \label{fig:sz-limit}
\end{figure}

Table \ref{tab:ceop} summarizes the outcomes achieved by GRASP and CRaSZe-AntS. In small-scale scenarios, such as setting the budget as $30\,\%$ of the best-known cost, GRASP outperforms CRaSZe-AntS. This is because GRASP focuses on optimizing a promising solution locally while ACS employs multiple ants to explore potential solutions across a broader scope. However, as problem instance scales increase, GRASP's `elite' mode diminishes. This is primarily due to a considerable increase in the time necessary to identify a promising solution and execute the `local search' operator. While for CRaSZe-AntS, the SZ layout significantly reduced the instance complexity, facilitating ACS to obtain a high-quality solution swiftly. For instance, when considering a $90\,\%$ budget allocation in \textit{bubbles7}, which comprises $406$ circles, GRASP took around $2.57$ times longer to obtain the solution than CRaSZe-AntS. While in \textit{bubbles8}, which involves 90 more circles, GRASP demonstrated a nonlinear time increase ($\sim$$3.95$ times longer) to obtain the solution over CRaSZe-AntS. Furthermore, CRaSZe-AntS performs more robustly compared to GRASP. For instance, when considering a $60\,\%$ budget level in \textit{bubbles9}, GRASP has a SD of 131.5 seconds for computation time and 99.28 for prize collection across 20 individual executions. Note that in \textit{bubbles9}, the prize for targets has a range of $[1, 12]$, this magnitude of $\mathscr{P}$ SD may represent more than eight targets unvisited in the solution path. While for CRaSZe-AntS, SDs associated with prize collection remain consistently limited in magnitude, specifically staying below 15 across all instances. On the contrary, this finding indicates that the SZ strategy narrows the range of path cost, leading to a solution with good quality but limited optimality - the SZ strategy increases the lower bound of the solution quality while potentially reducing its upper bound as well. This is more evident in CETSP scenarios, taking the CETSP instance \textit{bubbles6} shown in Fig. \ref{fig:sz-limit} as an example. \textit{bubbles6} features extreme scenarios, such as four circles intersecting at a single point. The investigation of such extreme scenarios falls beyond the scope of this paper because this `boundary problem' is typically associated with real-world constraints, such as data precision from GPS triangulation or sensor readings. Additionally, Fig. \ref{fig:overlap-in-ceop} visualizes the solution paths of GRASP and CRaSZe-AntS for \textit{kroD100} \citep{mennell2009heuristics} with $\phi_{or} = 0.02$ and \textit{dsj1000} \citep{mennell2009heuristics} with $\phi_{or} = 0.02$, respectively. CRaSZe-AntS attained solutions with higher prizes in both instances.

\subsection{Overlap of Non-uniform neighborhoods}
The advantage of considering overlap is further highlighted when dealing with non-uniform neighborhoods. This section presents solutions obtained through the BE scheme \citep{yang2018double} (in conjunction with PSO and IACS) and CRaSZe-AntS. 

\subsubsection{Algorithm parameter setting} \label{sec:TDDP-param}
We first demonstrate the limitations of the parameter settings recommended in \citep{yang2018double} for addressing the truck-and-drone delivery problem. \citet{yang2018double} suggested that setting the acceleration constants to $C_1 = C_2 = 2$ is generally effective. However, in the context of overlapped neighborhoods, such settings can lead to a phenomenon termed `velocity explosion' due to the limited search space in SZs. While it is possible to confine the waypoint to the SZ's boundary, large acceleration constants can increase the likelihood of the waypoint wandering around the boundary, thereby defeating the purpose of an interior search. To mitigate this issue, we opted for a more conservative setting $C_1 = C_2 = 1.33$, which reduces the probability of a particle overflying its momentary target to $\sim$25\,\% \citep{marini2015particle}. Additionally, to prevent the PSO from converging prematurely, we increase the minimum inertia to $\omega_{\min} = 0.4$ while maintaining the maximum inertia at $\omega_{\max} = 0.9$. It is worth noting that having a large number of particles (more than 50) can lead to unnecessary and premature computational efforts, as stated in \citep{marini2015particle}. Large population size may also conflict with the early stopping strategy, given the increased probability of constructing a distinct solution that outperforms others, particularly within the context of stochastic search. Therefore, we set the number of particles as $N_{\text{ptcl}} = 40$ and iterations $N_{\text{iter}}^{\text{PSO}} = 100$. Additionally, we set the minimum improvement tolerance as $\varepsilon^{\text{PSO}} = 10^{-4}$, so PSO would halt if the fitness difference $< 10^{-4}$ lasts for several iterations. Considering the computational time may be long in large-scale instances because PSO is coupled with ACS, we reduce the maximum iterations allowable without improvements. Specifically, we set this limit for PSO and IACS to $N_{\text{impr}}^{\text{PSO}} / 20 = 5$ and $N_{\text{impr}}^{\text{IACS}} = 13$. Though $N_{\text{impr}}^{\text{PSO}}$ and $N_{\text{impr}}^{\text{IACS}}$ are reduced, we employ the same PSO and IACS solver for both benchmark algorithm and CRaSZe-AntS. This should ensure a fair comparison between the two algorithms. Other settings remain the same as the ACS settings in \S \ref{sec:acs-param}.

\subsubsection{IACS convergence analysis}
To understand the convergence speed of IACS, we conducted a visual analysis of the first 20 iterations of PSO in solving a relatively large instance \textit{team5\_499} \citep{mennell2009heuristics}. Given that the optimization process of path prize may not show a distinct difference, we selected the path cost as the metric to represent convergence. We set $N_{\text{ptcl}} = N_{\text{ant}} = 20$, $N_{\text{iter}}^{\text{PSO}} = 20$, and $N_{\text{iter}}^{\text{ACS}} = N_{\text{iter}}^{\text{IACS}} = 100$ because solution quality is not the main focus in this experiment. Fig. \ref{fig:arc-search} illustrates the convergence dynamics of IACS and the classic ACS within three randomly selected particles of the PSO swarm. Compared to classic ACS, IACS can have a lower starting cost and a faster convergence across most iterations.
\begin{figure}[!t] 
    \centering
    \includegraphics[width=0.95\textwidth]{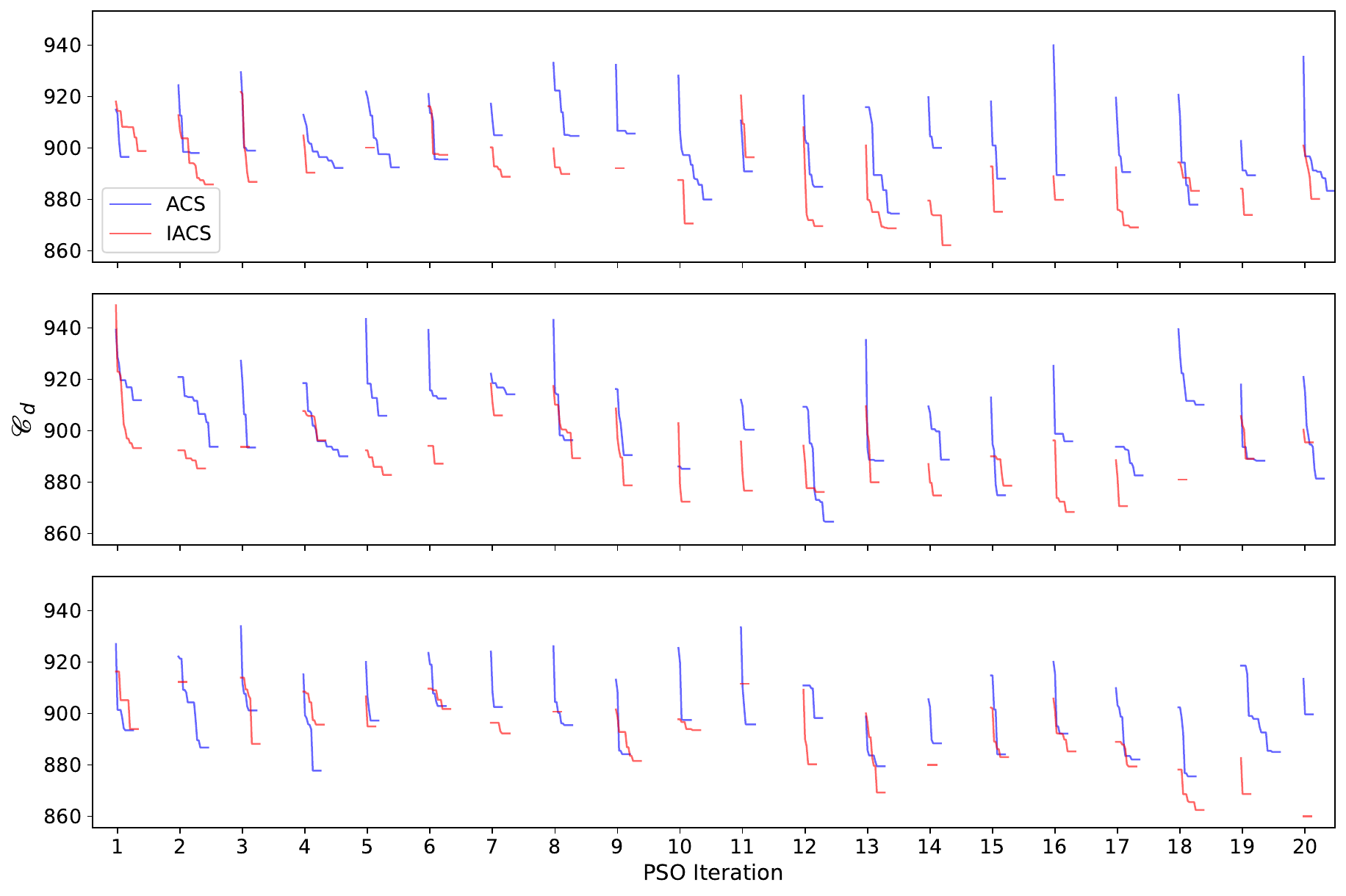}
    \caption{Convergence processes of ACS and IACS in the first 20 iterations of PSO when solving \textit{team5\_499}. Using the information from the last global-best ants, IACS can generally have a lower starting cost (except the first iteration).}
    \label{fig:iacs-convergence}
\end{figure}

\begin{table*}[!t]
\setlength\tabcolsep{0pt}
\setlength\extrarowheight{2pt}
\begin{threeparttable}[b]
\caption{Performance summary of BE-PSO-IACS and CRaSZe-AntS for TDDP} \label{tab:tddp}
\begin{tabular*}{\columnwidth}{@{\extracolsep{\fill}}*{14}{l}}
\toprule
\multirow{3}{*}{Instances} & \multirow{3}{*}{$\mathscr{B}_t$} & \multicolumn{6}{l}{BE-PSO-IACS} & \multicolumn{6}{l}{CRaSZe-AntS (RSZD-PSO-IACS)} \\ \cline{3-8} \cline{9-14} 
 &  & \multicolumn{2}{l}{$\mathscr{t}_\text{alg}$ (seconds)} & \multicolumn{2}{l}{$\mathscr{C}_t$} & \multicolumn{2}{l}{$\mathscr{P}$} & \multicolumn{2}{l}{$\mathscr{t}_\text{alg}$ (seconds)} & \multicolumn{2}{l}{$\mathscr{C}_t$} & \multicolumn{2}{l}{$\mathscr{P}$} \\ \cline{3-4}\cline{5-6}\cline{7-8} \cline{9-10}\cline{11-12}\cline{13-14}
  &  & AVG & SD & AVG & SD & AVG & SD & AVG & SD & AVG & SD & AVG & SD \\ \midrule
\textit{bubbles1} & 6.98 & 188.09 & 48.08 & 6.72 & 0.2 & 10.44 & 0.44 & 17.37 & 2.49 & 6.78 & 0.02 & \textbf{19.8} & 0 \\
\textit{bubbles2} & 8.57 & 276.01 & 72.83 & 8.45 & 0.07 & 9.38 & 0.22 & 42.83 & 12.09 & 8.44 & 0.07 & \textbf{21.82} & 0.86 \\
\textit{bubbles3} & 10.6 & 369.18 & 138.59 & 8.49 & 1.16 & 10.18 & 1.3 & 163.29 & 42.36 & 10.48 & 0.08 & \textbf{33.03} & 0.16 \\
\textit{bubbles4} & 16.11 & 591.83 & 1.41 & 16.02 & 0.06 & 19.8 & 0.34 & 539.28 & 63.46 & 15.99 & 0.11 & \textbf{52.33} & 1.24 \\
\textit{bubbles5} & 20.76 & 595.14 & 2.5 & 20.67 & 0.08 & 25.87 & 0.35 & 596.8 & 1.09 & 20.67 & 0.07 & \textbf{65.41} & 0.48 \\
\textit{bubbles6} & 24.59 & 598.98 & 8.23 & 24.49 & 0.09 & 30.09 & 0.36 & 597.15 & 1.41 & 24.51 & 0.07 & \textbf{73.67} & 0.46 \\
\textit{bubbles7} & 32.15 & 734.99 & 23.02 & 32.07 & 0.07 & 40.56 & 0.35 & 598.47 & 2.81 & 32.05 & 0.07 & \textbf{93.06} & 0.36 \\
\textit{bubbles8} & 38.93 & 1732.97 & 158.43 & 38.82 & 0.09 & 49.44 & 0.28 & 603.3 & 4.11 & 38.83 & 0.08 & \textbf{113.39} & 0.69 \\
\textit{bubbles9} & 45.18 & 2813.15 & 228.54 & 45.07 & 0.1 & 61.31 & 0.34 & 638.53 & 41.32 & 45.07 & 0.09 & \textbf{155.05} & 1.18 \\ \midrule
\textit{bubbles1} & 5.24 & 68.4 & 14.05 & 5.08 & 0.04 & 8.1 & 0 & 16.96 & 1.83 & 5.14 & 0.02 & \textbf{14.4} & 0 \\
\textit{bubbles2} & 6.42 & 141.42 & 46.45 & 6.26 & 0.06 & 8.22 & 0.04 & 31.38 & 8.05 & 6.28 & 0.09 & \textbf{15.86} & 0.31 \\
\textit{bubbles3} & 7.95 & 251.09 & 80.9 & 5.85 & 1.12 & 8.27 & 0.58 & 128.87 & 38.35 & 7.84 & 0.09 & \textbf{22.75} & 0.68 \\
\textit{bubbles4} & 12.08 & 591.08 & 0.89 & 12.03 & 0.06 & 13.59 & 0.25 & 462.7 & 120.56 & 12.02 & 0.06 & \textbf{39.67} & 0.51 \\
\textit{bubbles5} & 15.57 & 592.54 & 1.57 & 15.46 & 0.09 & 17.65 & 0.39 & 590.81 & 17.79 & 15.49 & 0.07 & \textbf{50.56} & 0.9 \\
\textit{bubbles6} & 18.44 & 595.34 & 3.36 & 18.35 & 0.07 & 20.84 & 0.45 & 596.34 & 0.76 & 18.34 & 0.06 & \textbf{56.2} & 1.44 \\
\textit{bubbles7} & 24.11 & 596.14 & 3.18 & 23.98 & 0.09 & 29.44 & 0.36 & 597.56 & 1.3 & 24.02 & 0.07 & \textbf{74.92} & 0.54 \\
\textit{bubbles8} & 29.2 & 744.29 & 35.67 & 29.09 & 0.09 & 36.7 & 0.33 & 599.33 & 3.06 & 29.12 & 0.06 & \textbf{90.92} & 0.58 \\
\textit{bubbles9} & 33.89 & 2068.68 & 178.73 & 33.78 & 0.07 & 45.13 & 0.5 & 599.84 & 3.78 & 33.74 & 0.13 & \textbf{118.07} & 1.03 \\ \midrule
\textit{bubbles1} & 3.49 & 29.05 & 4.88 & 3.23 & 0.04 & 4.5 & 0 & 7.14 & 0.78 & 3.27 & 0.01 & \textbf{8.1} & 0 \\
\textit{bubbles2} & 4.28 & 86.07 & 17.21 & 4.19 & 0.08 & 5.98 & 0.4 & 13.87 & 3.88 & 4.19 & 0.07 & \textbf{9.82} & 0.55 \\
\textit{bubbles3} & 5.3 & 205.73 & 76.6 & 5.13 & 0.07 & 7.89 & 0.26 & 51.29 & 4.82 & 5.21 & 0.02 & \textbf{16.27} & 0 \\
\textit{bubbles4} & 8.05 & 580.96 & 25.51 & 7.95 & 0.06 & 10.08 & 0.2 & 282.78 & 104.28 & 7.95 & 0.08 & \textbf{25.45} & 0.75 \\
\textit{bubbles5} & 10.38 & 591.61 & 1.28 & 10.28 & 0.06 & 10.98 & 0.27 & 439.05 & 107.1 & 10.28 & 0.08 & \textbf{33.79} & 0.53 \\
\textit{bubbles6} & 12.3 & 592.16 & 1.72 & 12.18 & 0.09 & 12.73 & 0.4 & 492.25 & 91.07 & 12.22 & 0.08 & \textbf{36.9} & 1.41 \\
\textit{bubbles7} & 16.07 & 592.58 & 2.11 & 15.97 & 0.08 & 17.51 & 0.37 & 596.43 & 0.81 & 15.99 & 0.06 & \textbf{52.07} & 0.91 \\
\textit{bubbles8} & 19.47 & 593.72 & 3.47 & 19.37 & 0.09 & 22.43 & 0.33 & 598.15 & 1.9 & 19.35 & 0.09 & \textbf{63.21} & 0.72 \\
\textit{bubbles9} & 22.59 & 862.47 & 45.03 & 22.49 & 0.07 & 27.26 & 0.42 & 598.72 & 2.27 & 22.51 & 0.06 & \textbf{76.95} & 0.68 \\
\bottomrule
\end{tabular*}
\scriptsize
\vspace*{0.1cm}
\parbox[b]{\textwidth}{CRaSZe-AntS outperforms BE-PSO-IACS across all instances tested, with an averaged $\mathscr{t}_{\text{alg}}$ reduction of $-55.18\,\%$ and $\mathscr{P}$ increase of $140.44\,\%$.}
\end{threeparttable}
\end{table*}

\subsubsection{The performance of CRaSZe-AntS in TDDP instances}
Following \citet{chang2018optimal}, we set parameters of the TDDP model (as stated in \S \ref{sec:TDDP}) as follows: the UAV average flight speed $\mathcal{V}_{\text{drone}} = 90$ km/h; the UAV average service time $t_{\text{serv}} = 5$ minutes; the maximum service range for the UAV $r_{\text{drone}} = 10$ km; the number of UAVs deployed on a truck $N_{\max}^{|\Omega|} = 5$; the truck average driving speed $\mathcal{V}_{\text{truck}} = 60$ km/h; UAV flight efficiency bounds $\lambda_{\min} = 80\,\%, \: \lambda_{\max} = 100\,\%$. The original prizes assigned to targets in CEOP instances are normalized within the interval $[ 0.1, 0.9 ]$. This adjustment aims to prevent the pheromone levels from being excessively elevated. We converted distance budgets $\mathscr{B}_d$ into time budgets $\mathscr{B}_t$ by dividing the best-known distance costs by the truck's average speed. Given that TDDP entails additional time costs associated with prize collection, we set budgets at $60\,\%, 90\,\%$, and $120\,\%$ levels. Note that we set a higher budget level for TDDP instances because here $\mathscr{B}_t = \text{best-known CETSP cost}\: /\: \mathcal{V}_{\text{truck}}$, which ignores the truck idle time.

\begin{figure*}[!b] 
    \centering
    \begin{subfigure}{\textwidth} 
        \centering
        \includegraphics[width=\textwidth]{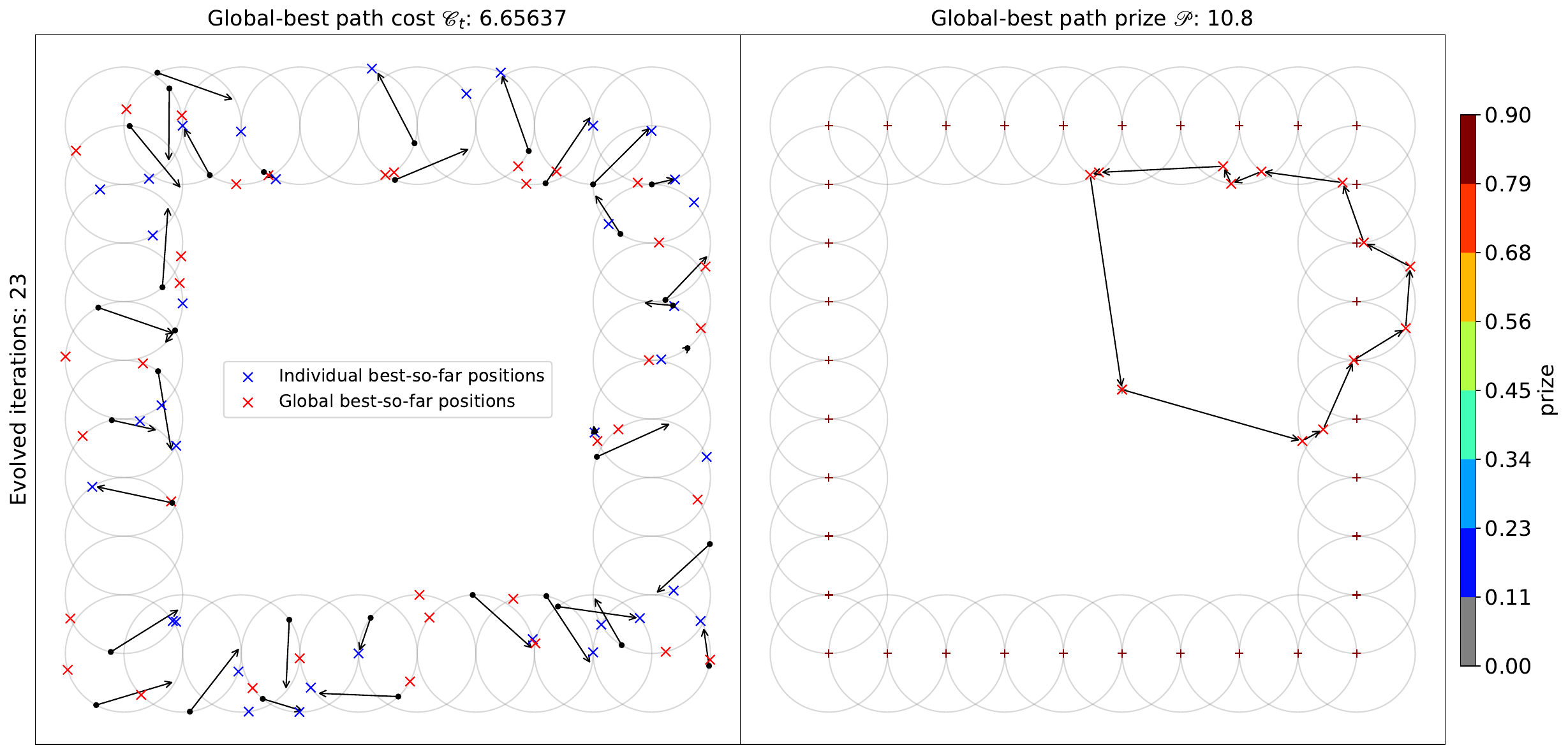}
        \caption{Solution path generated by BE-PSO-IACS. The left sub-figure refers to the waypoint positions of a randomly selected particle at the last iteration (before break). Some waypoints converged the global best-so-far positions. The right sub-figure shows the global best-so-far path with a corresponding color bar for circles' prizes.} \label{fig:BE-PSO-ACS-TDDP}   
    \end{subfigure}
    \begin{subfigure}{\textwidth} 
        \centering
        \includegraphics[width=\textwidth]{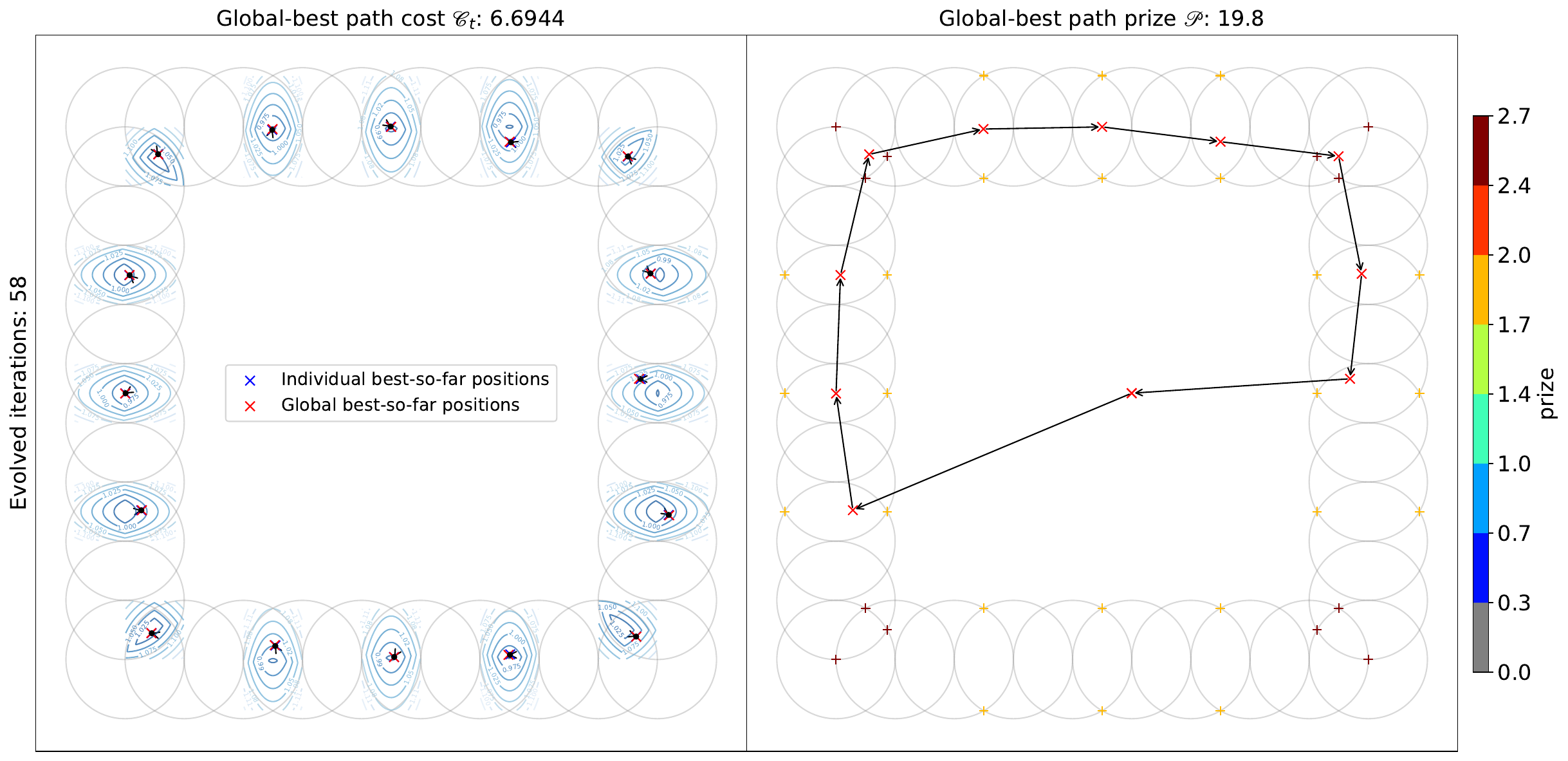}
        \caption{Solution path generated by CRaSZe-AntS. The left sub-figure refers to the waypoint positions of a randomly selected particle at the last iteration (before break). Most waypoints converged to the global best-so-far positions. The blue contour lines represent the time cost of the SZ. The right sub-figure shows the global best-so-far path with a corresponding color bar for SZs' prizes.} \label{fig:RSZD-PSO-ACS-TDDP}
    \end{subfigure}
    \caption{Visualization of BE-PSO-IACS's and CRaSZe-AntS's solution in solving TDDP instance \textit{bubbles1}, with given budget $\mathscr{B}_t = 6.9826$.}
    \label{fig:TDDP}
\end{figure*}

In our comparative analysis, the Hybrid Algorithm (HA) proposed by \citet{yang2018double} emerges as the most analogous algorithm to CRaSZe-AntS for solving our TDDP. HA incorporated PSO with the LDIW strategy for continuous waypoint position optimization and GA with partial mapping crossover for discrete path sequence optimization. \citet{yang2018double} proposed the Boundary Encoding (BE) scheme for arbitrary neighborhoods, simplifying the two-dimensional waypoint coordinates into polar angles, allowing velocity vectors to be confined within the range $\mathcal{V}_j \in [-\pi, \pi]$. In this experiment, we still adopted the BE scheme during PSO optimization while replacing GA with IACS for sequence optimization. This is because a) some parameter configurations of original PSO are inapplicable under TDDP scenarios, as stated in \S \ref{sec:TDDP-param}, which means direct comparison between original HA and CRaSZe-AntS would be unfair; and b) any disparities between GA and IACS can be mitigated. Table~\ref{tab:tddp} presents the outcomes of BE-PSO-IACS and CRaSZe-AntS for \textit{bubbles1-9}. An early termination mechanism has been implemented, whereby the system monitors the cumulative execution time before updating the velocity of individual particles. In other words, the maximum allowed execution time is 10 minutes due to the extended computation time for 20 executions of every TDDP instance. Our experimental results reveal marked distinctions between BE-PSO-IACS and CRaSZe-AntS. For instance, CRaSZe-AntS significantly reduces computation time and delivers prizes approximately twice as high as those achieved by BE-PSO-IACS (an increasing gap can be observed). Both CRaSZe-AntS and BE-PSO-IACS demonstrate stable performances across all executions. Note that in some large-scale instances, such as \textit{bubbles8-9} with $120\,\%$ budget level, CRaSZe-AntS and BE-PSO-IACS fail to converge within the time limit. This is because the time needed for their initialization phase is longer than 10 minutes. Fig. \ref{fig:TDDP} demonstrates the waypoint positions of a randomly selected particle at the last iteration and the global-best solutions generated from BE-PSO-IACS and CRaSZe-AntS. Fig. \ref{fig:BE-PSO-ACS-TDDP} reveals that less than half of the waypoints converged to the global best-so-far positions because of the large difference between the individual and global best-so-far solutions. This observation indicates the inefficiency of boundary-searching techniques when applied within the context of TDDP. Furthermore, as can be seen in the right part of Fig. \ref{fig:RSZD-PSO-ACS-TDDP}, the final solution path demonstrates a good balance between accommodating the truck's travel time and waiting time within the budget constraint.

\subsection{Complexity Analysis} \label{sec:complexity}
Let the number of target circles $N$ be the primary input. For CEOP, CRaSZe-AntS has a worst-case time complexity of $\mathcal{O}(N^2)$ because the most computationally demanding component is the 2-opt operator, which has a complexity of $\mathcal{O}(N^2)$. On the other hand, the benchmark algorithm, GRASP, is an iterative approach with a \textbf{base} time complexity of $\mathcal{O}(N^2)$. Under certain conditions, such as when GRASP identifies a high-quality solution that GRASP's local search operator cannot readily improve, the algorithm may terminate prematurely after only a few executions of the 2-opt operator. However, maintaining a time complexity of $\mathcal{O}(N^2)$ cannot be guaranteed for GRASP in practice, particularly for large-scale instances. 

For TDDP, CRaSZe-AntS has a worst-case time complexity of $\mathcal{O}(N^2)$ because RSZD has a complexity of $\mathcal{O}(N_{\text{iter}}^{\text{RSZD}}\cdot N_{\max}^{|\Omega|} \cdot N)$ under a constraint of maximum circles, the solution construct phase of PSO and IACS has a complexity of $\mathcal{O}(N_{\text{iter}}^{\text{PSO}} \cdot N_{\text{ptcl}} \cdot N_{\text{iter}}^\text{{IACS}} \cdot N_\text{{ant}} \cdot N)$ and 2-opt local search can have a complexity of $\mathcal{O}(N^2)$. BE-PSO-IACS has the same worst-case complexity of $\mathcal{O}(N^2)$. Though the benchmark algorithms have the same theoretical complexity as our algorithms for addressing both CEOP and TDDP, the actual computation time of CRaSZe-AntS is significantly reduced by SZs.

\section{Limitation and Future Work}
As the primary focus of this work is to explore the benefits of considering overlaps in solving CEOP and CEOP-$\mathcal{N}$ rather than to generate a high-quality SZ layout, the RSZD scheme may benefit from minimum-angle strategy and decomposition of high-degree SZs, as presented in \citep{mennell2009heuristics}. Further enhancement can be achieved through incorporating the boundary encoding scheme \citep{yang2018double}. This scheme facilitates the extension of current SZs into generalized Steiner Zones, i.e., the overlapped regions formed by arbitrary convex neighborhoods. Moreover, a comprehensive sensitivity analysis of CRaSZe-AntS, including understanding the impact of RSZD iterations on \textit{bubbles} series and RSZD's performance on more cluttered instances, is left for future work. Additionally, while our PSO and ACS solver involve strategies including maximum velocity constraint and inheritance mechanism to improve their efficiency, embedding ACS within PSO may be computationally expensive in large-scale CEOP-$\mathcal{N}$ instances. Therefore, considering alternative approaches, such as replacing either PSO or ACS with machine learning techniques like attention mechanisms and reinforcement learning \citep{kool2018attention} may offer computational latency improvements. Moreover, the truck-and-drone delivery model in this paper only considers necessary constraints. A more comprehensive simulation of real-world scenarios may require additional details, such as accounting for daily weather conditions as a factor affecting UAV flight speed. A more accurate representation of UAV operations may also be considered instead of relying on idealized assumptions with constant time costs. Finally, involving three-dimensional coordinates and neighborhoods can further improve the performance of CEOP-$\mathcal{N}$, particularly for UAV mission planning problems.

CEOP-$\mathcal{N}$ offers practical advantages over alternative `close-enough' formulations, particularly when addressing `one-to-many' scenarios. While our investigation has primarily focused on the spatial aspect of the cost function, it can be extended to general continuous functions, thereby facilitating the development of a highly generalizable model. Many real-world applications can be modeled well with CEOP-$\mathcal{N}$, such as mass product distribution \citep{derya2020selective}, collection of medical waste \citep{mantzaras2017optimization}, rebalancing of shared bikes \citep{schuijbroek2017inventory}, wireless data collection with UAV \cite{chen2021data}, etc. Our future work will concentrate on enhancing the efficiency of our algorithm and applying it to more complicated real-world UAV mission planning problems that may involve wireless data collection tasks. Thus, three-dimensional generalized Steiner Zones will be explored within this context. Additionally, as stated in \S \ref{sec:complexity}, the dual metaheuristics framework may exhibit unacceptable computational time in some scenarios, particularly when the computing resource is constrained, such as on the onboard computer. Such systems typically require lightweight online algorithms due to dynamic environmental conditions. Moreover, the Branch-and-Bound (BnB) algorithm introduced by \cite{coutinho2016branch} sets an important benchmark for evaluating algorithms designed to solve the CETSP. Adapting exact methods like the BnB algorithm to CEOP and CEOP-$\mathcal{N}$, however, introduces considerable challenges. Unlike CETSP, where the number of target circles predetermines the solution's length, CEOP and CEOP-$\mathcal{N}$ may require multiple iterations to determine the optimal solution length. Given these intricate challenges, research and development of exact methodologies for CEOP and CEOP-$\mathcal{N}$ remains an open challenge that could offer important complementary perspectives to existing heuristics.

\section{Conclusion} \label{sec:conclusion}
Although the Close Enough Orienteering Problem (CEOP) has been receiving attention within the research community, few studies explore the interior of the target neighborhood and discuss the potential impact of overlapped neighborhoods within this context. This paper introduces an efficient Steiner Zone (SZ) based strategy, named CRaSZe-AntS, for addressing the CEOP, specifically emphasizing handling uniform and non-uniform cost functions for neighborhood prize collection. CRaSZe-AntS swiftly constructs a valid SZ layout and identifies SZs' vertices by the proposed Randomized Steiner Zone Discretization (RSZD) scheme. The (near-)optimal positions of waypoints can then be determined quickly by the arc search algorithm based on the geometry of constructed SZs and their vertices. This information can be further utilized for scenarios of non-uniform neighborhoods to define the boundary condition of continuous interior search offered by the Particle Swarm Optimization (PSO) algorithm and maximum velocity constraints in the PSO. Additionally, our solver for determining the sequence order of waypoints, i.e., Ant Colony System (ACS), employs a new special inheritance mechanism. IACS utilizes the prior knowledge between consecutive iterations, which is empirically proven to be generally faster than classic ACS under this scenario. Our experimental results demonstrate that CRaSZe-AntS is computationally efficient and reliable in addressing both CEOP and CEOP-$\mathcal{N}$, compared to state-of-the-art single-neighborhood-based strategies.

\section*{Acknowledgments}
We thank Prof. Bruce L. Golden for providing the CETSP instances in \citep{mennell2009heuristics}.

\bibliographystyle{cas-model2-names}
\bibliography{references}

\newpage
\appendix
\noindent {\textbf{\Large Appendix}} 
\vspace*{-0.2cm}
\section{Graphical illustration of CRaSZe-AntS architecture}
\begin{figure}[!htbp]
    \centering
    \includegraphics[width=0.5\textwidth]{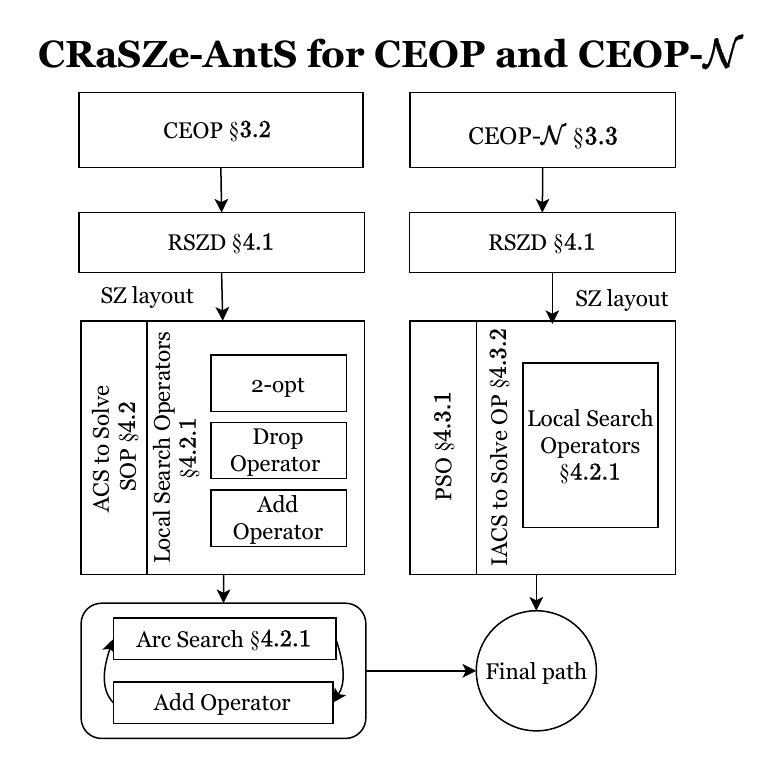}
    \caption{Graphical illustration of CRaSZe-AntS architecture} \label{fig:graphical-abstract}
\end{figure}

The overall architecture of CRaSZe-AntS for CEOP and CEOP-$\mathcal{N}$ can be seen in Fig. \ref{fig:graphical-abstract}. The detailed workflow is available online via the GitHub link \url{https://github.com/sysal-bruce-publication/CRaSZe-AntS}.

\section{Formulation for the Set Orienteering Problem (SOP)} \label{sec-appendix}
In alignment with the notation used by \citet{pvenivcka2019variable}, the decision variables are:
\begin{itemize}
    \item[$Y_i$\: :] binary variable equals to 1 if a vertex within the SZ $\Omega_i$ is visited, and 0 otherwise;
    \item[$X_{ij}$\: :] binary variable equals to 1 if edge $e_{ij}$ is traversed, and 0 otherwise.
\end{itemize}
We formulate our SOP as an Integer Linear Programming (ILP) problem based on \citep{pvenivcka2019variable}: 
\begin{subequations}
\begin{flalign}
    \label{eq:sop-obj}
    \textbf{(SOP)} & \quad\max\sum_{\Omega_i \in\: \mathbf{Z}} \mathscr{P}(\Omega_i) \cdot Y_i &\\
    \label{eq:sop-budget}
    \textbf{s.t.}\quad & \sum_{e_{ij} \in\: \mathbf{E}} \mathscr{C}_d \big( v'_i, v'_j \big) \cdot X_{ij} \leq \mathscr{B}_d &\\
    \label{eq:sop-same-edge}
    & \sum_{v'_i \in\: \mathbf{V}\setminus \{ \Omega_k \}} X_{ij} = \sum_{v'_i \in\: \mathbf{V}\setminus \{ \Omega_k \}} X_{ji} & \forall\: \Omega_k\in \mathbf{Z},\; \forall\: v'_j \in \Omega_k &\\
    \label{eq:sop-visited-sz-edge-entering}
    & \sum_{v'_i\in\: \mathbf{V}\setminus\{\Omega_k\}} \: \sum_{v'_j\in \Omega_k} X_{ij} = Y_{k} & \forall\: \Omega_k\in \mathbf{Z} &\\ 
    \label{eq:sop-visited-sz-edge-leaving}
    & \sum_{v'_i\in\: \mathbf{V}\setminus\{\Omega_k\}} \: \sum_{v'_j\in \Omega_k} X_{ji} = Y_{k} & \forall\: \Omega_k\in \mathbf{Z} &\\
    \label{eq:sop-subtour}
    & \sum_{v'_i \in\: \mathbf{S}} \: \sum_{v'_j \in\: \mathbf{S}} X_{ij} \leq \sum_{\Omega_k\in\: \mathbf{S}\setminus\{\Omega_l\}} Y_k, & \forall\: \mathbf{S}\subset \mathbf{Z},\: \forall\: \Omega_l \in \mathbf{S} &\\
    \label{eq:sop-start-end}
    & \sum_{j=1}^{K+1} X_{0j} = \sum_{i=0}^{K} X_{i K+1} = 1 &
\end{flalign}
\end{subequations}
The objective function \eqref{eq:sop-obj} maximizes the collected prizes from SZs. Constraint \eqref{eq:sop-budget} implies the path distance cost must not exceed the given distance budget $\mathscr{B}_d$. Constraint \eqref{eq:sop-same-edge} ensures each SZ vertex has the same number of entering and leaving edges. Constraint \eqref{eq:sop-visited-sz-edge-entering} and \eqref{eq:sop-visited-sz-edge-leaving} enforces each visited SZ has exactly one entering and one leaving edge, otherwise no entering and leaving edge. Constraint \eqref{eq:sop-subtour} eliminates subtours. Constraint \eqref{eq:sop-start-end} requires the path starts at $v'_0$ and ends at $v'_{K+1}$.

\end{document}